\pdfoutput=1
\documentclass[11pt]{article}

\usepackage{authblk}
\usepackage[]{ACL2023}
\usepackage{lipsum}
\usepackage{multirow}
\usepackage{times}
\usepackage{latexsym}
\usepackage[normalem]{ulem}
\useunder{\uline}{\ul}{}
\usepackage{amsmath}
\usepackage{amssymb}
\usepackage{array}
\usepackage{latexsym}
\usepackage{algorithm,algorithmic}
\usepackage{babel}
\usepackage{graphicx}
\usepackage{xcolor,colortbl}
\usepackage{cleveref}

\usepackage[T1]{fontenc}

\newcolumntype{P}[1]{>{\centering\arraybackslash}p{#1}}
\newcommand{\method}{\texttt{CT-CBM}}
\usepackage[utf8]{inputenc}
\usepackage{bbm}
\usepackage{microtype}

\usepackage{inconsolata}

\title{Towards Achieving Concept Completeness\\ for Textual Concept Bottleneck Models}

\author[1,2]{Milan Bhan$^*$}
\author[2]{Yann Choho$^*$}
\author[2]{Pierre Moreau}
\author[1]{Jean-Noël Vittaut}
\author[2]{Nicolas Chesneau}
\author[1]{Marie-Jeanne Lesot}
\affil[1]{Sorbonne Université, CNRS, LIP6, F-75005 Paris, France}
\affil[2]{Ekimetrics, Paris, France}
\affil[ ]{\texttt{\{milan.bhan, yann.choho, pierre.moreau, nicolas.chesneau\}@ekimetrics.com}}
\affil[ ]{\texttt{\{jean-noel.vittaut, marie-jeanne.lesot\}@lip6.fr}}

\begin{document}
\maketitle
\def\thefootnote{*}\footnotetext{These authors contributed equally to this work.}
\begin{abstract}
Textual Concept Bottleneck Models (TCBMs) are interpretable-by-design models for text classification that predict a set of salient concepts before making the final prediction. 
%However, existing TCBM face significant limitations, including reliance on computationally expensive large language models, dependence on predefined human-labeled concepts, vulnerability to classification leakage, and unreliable concept detection. 
This paper proposes Complete Textual Concept Bottleneck Model (\method), a novel TCBM generator building concept labels in a fully unsupervised manner using a small language model, eliminating both the need for predefined human labeled concepts and LLM annotations. \method\ iteratively targets and adds important and identifiable concepts in the bottleneck layer to create a complete concept basis. 
\method\ achieves striking results against competitors in terms of concept basis completeness and concept detection accuracy, offering a promising solution to reliably enhance interpretability of NLP classifiers.
\end{abstract}

\section{Introduction} 
\label{intro}

The striking level of performance in natural language processing (NLP) achieved by black-box neural language models~\cite{vaswani_attention_2017, gpt3, chowdhery2023palm} comes along with a lack of interpretability~\cite{madsen2022survey}. The field of eXplainable Artificial Intelligence (XAI)~\cite{xai_2_0_survey} intends to make the behavior of such models more interpretable. A common distinction of XAI is to define interpretability methods either (1) by applying post hoc explanation methods to interpret black box models,  or (2) by constructing interpretable models by-design~\cite{jacovi2020towards, madsen2024interpretability}. 

\begin{table}[t]
\small
\centering
\begin{tabular}{cccc}
\hline
\rowcolor[HTML]{ECF4FF} 
                                                                         & \textbf{\texttt{C3M}} & \textbf{\texttt{CB-LLM}} & \textbf{\method} \\
\rowcolor[HTML]{ECF4FF} 
                                                                         &                       &                          & \textbf{(ours)}  \\ \hline
\rowcolor[HTML]{FFFFFF} 
\begin{tabular}[c]{@{}c@{}}Need for predefined\\  concepts\end{tabular}      & Yes                   & \textbf{No}              & \textbf{No}      \\ \hline
\rowcolor[HTML]{EFEFEF} 
Use of LLM                                                               & Yes                   & Yes                      & \textbf{No}      \\ \hline
\rowcolor[HTML]{FFFFFF} 
Scalability                                                              & No                    & \textbf{Yes}             & \textbf{Yes}     \\ \hline
\rowcolor[HTML]{EFEFEF} 
\begin{tabular}[c]{@{}c@{}}Black-box \\ performance reached\end{tabular} & \textbf{Yes}          & \textbf{Yes}             & \textbf{Yes}     \\ \hline
\rowcolor[HTML]{FFFFFF} 
\begin{tabular}[c]{@{}c@{}}Concept base\\ completeness\end{tabular}      & No                    & No                       & \textbf{Yes}     \\ \hline
\rowcolor[HTML]{EFEFEF} 
\begin{tabular}[c]{@{}c@{}}Accurate concept\\ detection\end{tabular}     & No                    & No                       & \textbf{Yes}     \\ \hline
\end{tabular}
\caption{\label{tab:intro} 
Qualitative comparison of \method\ to competitors. Desired modalities are highlighted in bold.}
\end{table}

One promising approach in the second category is Concept Bottleneck Models (CBM)~\cite{cbm_intro}. CBM are models that first map the input representations to a set of human-interpretable high-level attributes, called \emph{concepts}, in a Concept Bottleneck Layer (CBL). These concepts are then linearly projected to make the final prediction, improving the interpretability of black box models. While CBM have been widely used in computer vision~\cite{cbm_posthoc, label_free_cbm, cbm_incremental, concepf_embedding_models}, they have been much less explored for NLP~\cite{concept_xai_survey}. Existing Textual Concept Bottleneck Models (TCBM) have limitations: (i) they mainly rely on the use of large language models (LLM)~\cite{cbm_plm, cbm_crafting,  cbm_by_design} whose computational cost is prohibitive, (ii) they often require access to a set of predefined human-labeled concepts~\cite{cbm_plm, cbm_by_design}, (iii) the concept base of the CBL can be over complete, making the CBM predictions based on too many concepts and difficult to understand~\cite{cbm_plm, cbm_crafting} (iv) they do not systematically guarantee the reliability of the CBL concept detection, making the corresponding explanations unfaithful.

In this paper, we propose Complete Textual Concept Bottleneck Model (\method), a novel approach to transform any fine-tuned NLP classifier into an interpretable-by-design TCBM accurately detecting concept from a complete concept basis. As summarized in Table~\ref{tab:intro}, the main contributions of \method\ are as follows: \begin{enumerate}
    \item We present a computationally affordable method to perform concept discovery and annotation in a fully unsupervised manner, solely based on a small language model.
    \item We propose a new method to target relevant concepts to be added in the CBL, based on local concept importance and  identifiability.
    \item Concept completeness, estimated as the smallest concept base to cover the dataset while enabling the TCBM to approximately replicate the performance of the initial black-box model, is achieved through iterative addition of concepts in the CBL.
\end{enumerate}
This way, the \method\ method we propose offers both (1) an affordable concept annotation method to construct a concept bank and (2) a method to generate a TCBM derived from any initial concept bank, with an accurate CBL constructed upon a complete concept base. 

The paper is organized as follows:  Section~\ref{bk_rw} recalls some basic principles of XAI and related work. Section~\ref{method} describes the proposed \method. Section~\ref{xp} presents the conducted experiments, that show that \method\ systematically succeeds in reaching the performance of its competitors in terms of downstream task accuracy and detects significantly more precisely the concepts present in its concept layer, while containing fewer concepts than its competitors.

\section{Background and Related Work}
\label{bk_rw}
This section first recalls some principles of XAI methods used later in the papers and presents existing methods generating Concept Bottleneck Model in general, and for NLP.
% \begin{itemize}
%     \item Def concepts and TCAV
%     \item Def CBM
%     \item Strategies: projection, and training (joint, sequential and independant)
%     \item With and without label, iteratively, learning to intervene
%     \item No label: use of LLM (chatgpt...) and mainly for CV
% \end{itemize}

\subsection{XAI Background}
% \paragraph{Neural Post Hoc Interpretability.} 
\paragraph{Post Hoc Interpretability.}
Post hoc methods explain the behavior of a model after its training. They first include attribution methods that compute importance scores to identify the input dimensions that are mostly responsible for the obtained results, 
%~\cite{zhao_explainability_2023}, such as  the gradient-based approach 
e.g. 
\texttt{Integrated Gradients}~\cite{sundararajan_axiomatic_2017}. 
% to inputs to explain the model outcome~\cite{zhao_explainability_2023}. In particular, gradient-based approaches such as \texttt{Integrated Gradients}~\cite{sundararajan_axiomatic_2017} compute these scores by back-propagating the gradients through the model. 

Second, post hoc concept-based approaches generate explanations at a higher level of abstraction, by focusing on human interpretable attributes, called \emph{concepts}. Given a concept exemplified by some user defined examples, \texttt{TCAV}~\cite{tcav} assesses the model's sensitivity to the latter by back-propagating the gradients with respect to a linear representation of the considered concept, called concept activation vector (CAV).
% In the original paper, 
% \texttt{TCAV} relies on human-labeled concepts, whose annotation can be time-consuming and expensive.  

\paragraph{Concept Bottleneck Models.}
Another way to improve
the interpretability of AI systems consists in constructing so-called interpretable-by-design Concept Bottleneck Models (CBM)~\cite{cbm_intro}. They sequentially detect concepts in a Concept Bottleneck Layer (CBL) and linearly make the final prediction from the latter, thereby significantly improving the understanding of the decision-making process. CBM face several limitations: (1) they require predefined human-labeled concepts, (2) their CBL are often incomplete, leading to either reduced model accuracy (under-complete concept base) or unintelligible explanations (over-complete concept base)~\cite{cbm_incremental}, (3) they are vulnerable to downstream task leakage, where prediction models inadvertently use unintended signals from concept predictor scores rather than the actual concepts, compromising both the faithful detection of concepts and the overall interpretability of CBM~\cite{addressing_leakage_cbm}, (4) they do not ensure accurate concept prediction in the CBL.

% CBM have some limitations, such as requiring a predefined set of human-labeled concepts, generating incomplete concept bottleneck layers (CBL), or suffering from downstream task leakage. Concept incompleteness can have consequences either on the model accuracy (under-complete concept base) or the intelligibility (over-complete concept base) of the provided explanations~\cite{cbm_incremental}. Downstream task leakage~\cite{addressing_leakage_cbm} occurs when the final prediction uses unintended additional information from the concept predictor scores. The concept predictor then no longer needs to detect faithfully the concepts to be accurate on the classification task, thus compromising the interpretability faithfulness of the CBM. 

Among the extensive CBM literature~\cite{concept_survey}, several approaches have been proposed to address specific limitations. \texttt{Label-Free CBM}~\cite{label_free_cbm} employs GPT-3~\cite{gpt3} to identify key concepts for class recognition, eliminating the need for predefined concepts. Other approaches~\cite{cbm_posthoc, addressing_leakage_cbm} add a non-interpretable parallel residual connection to match black-box NLP classifier accuracy and mitigate leakage by processing unintended information through this residual layer, though reducing the interpretability of the CBM.
\texttt{Res-CBM}~\cite{cbm_incremental} derives new concepts from the residual layer to create more complete CBLs, 
%concept bottleneck layers, 
but still requires predefined concept candidates. While these methods overcome or mitigate certain CBM limitations, their application has primarily been restricted to computer vision.

% Among the vast literature on CBM~\cite{concept_survey}, numerous variants have been proposed to address one by one the aforementioned limitations. Notably, \texttt{Label-Free CBM}~\cite{label_free_cbm} prompts GPT-3~\cite{gpt3} to list the most important concepts for recognizing a specific class, freeing the approach from dependency on predefined labeled concepts. However, \texttt{Label-Free CBM} structurally depends on the parametric knowledge of GPT-3 and does not generate data driven concepts. In order to reach the accuracy of a black box NLP classifier while avoiding leakage, a non-interpretable connection parallel to the concept layer can be added to fit the residuals between the raw CBM outcome and the ground truth~\cite{cbm_posthoc, addressing_leakage_cbm}. However, adding such a residual connection decreases the CBM interpretability. \texttt{Res-CBM}~\cite{cbm_incremental} develops a method to derive new concepts from the residual layer to build a more complete CBL. Yet, it 
% %\texttt{Res-CBM}~\cite{cbm_incremental} 
% requires access to a set of candidate concepts to add to the CBL before probing the residual connection. Although these methods overcome some of the limitations inherent in CBM, their application has so far been restricted to computer vision. 

\subsection{Textual Concept Bottleneck Models}
\label{related_work}
This section presents recent works on generating Concept Bottleneck Models for NLP, referred to  as Textual Concept Bottleneck Models (TCBM).

\texttt{TBM}~\cite{cbm_by_design} iteratively discovers concepts by leveraging GPT-4~\cite{gpt4} and focusing on examples misclassified by a separately trained linear layer. \texttt{TBM} is not strictly a CBM, since concept detection is performed with GPT-4 during inference, making also the approach non scalable and computationally expensive.

\texttt{C\textsuperscript{3}M}~\cite{cbm_plm} supplements human-labeled concepts with ChatGPT-generated concepts, achieving performance comparable to black-box NLP classifiers. However, this approach builds a TCBM with an over-complete concept basis, and its dependence on ChatGPT and human labeling limits its reproducibility and scalability.

\texttt{CB-LLM}~\cite{cbm_crafting, cb_llm} also uses ChatGPT to generate concept candidates that are then scored using a sentence embedding model to create numerical concept representations, making the approach more affordable than \texttt{C\textsuperscript{3}M}~\cite{cbm_plm}. While matching black-box classifier performance, it also neglects CBL completeness, potentially leading to unreliable concept detection and unintelligible explanations.

\begin{figure*}[t]{\centering}
\begin{center}
\includegraphics[scale=0.50]{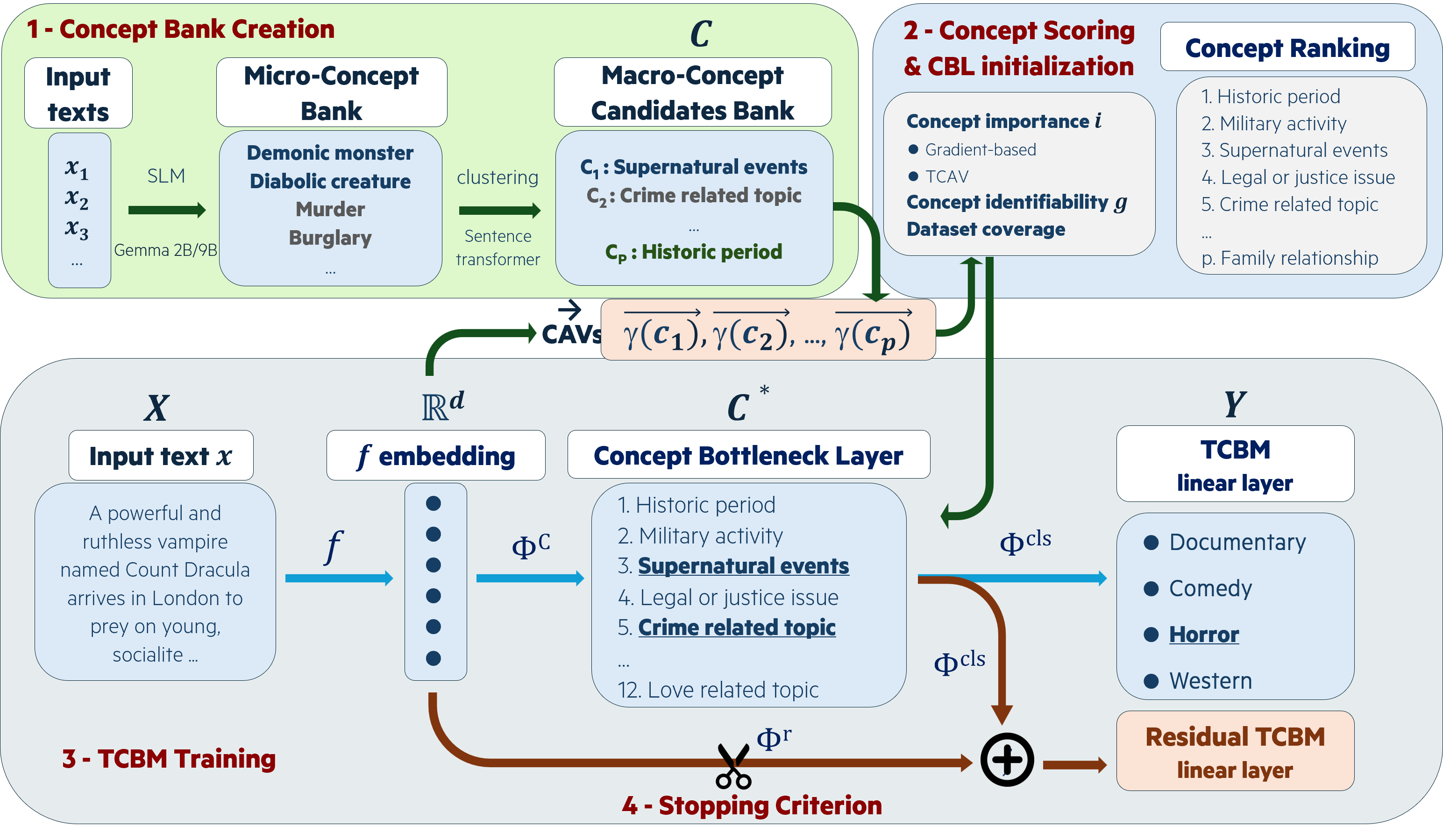}
\caption{\method\ overview illustrated with film synopsis classification. \method\ is a 4-step approach to build a TCBM from a $f$ black box NLP classifier. (1) A concept bank is created from the text corpus of interest. (2) Concepts are scored given their importance to explain $f$ predictions and their identifiability score, and the CBL is initialized. (3) The TCBM is trained through 3 layers: $\Phi^{\text{C}}$, $\Phi^{\text{cls}}$ and $\Phi^{\text{r}}$. (4) The TCBM training stops when the performance of the TCBM with $\Phi^{\text{r}}$ is reached without the latter; $\Phi^{\text{r}}$ is finally removed.}  
\label{fig:global_flow}
\end{center}
\end{figure*}

\section{Proposed approach: \method} 
\label{method}
This section describes our proposed Complete Textual Concept Bottleneck Model (\method). As shown in Table~\ref{tab:intro}, \method\ uses only small language models (SLM) for concept discovery, eliminating the need for predefined human-labeled concepts, and builds a TCBM with a complete concept bottleneck layer containing properly detected and relevant concepts. 

\subsection{\method\ Overview}
We consider a corpus of text-label pairs $\mathcal{T} = \mathcal{X} \times \mathcal{Y} = \left\{ (x,y)\right\}$ where $x \in \mathcal{X}$ denotes the text and ${y \in \mathcal{Y}}$ the label. We respectively denote $\mathcal{X}_{train}, \mathcal{X}_{dev}$ and $\mathcal{X}_{test}$ the training, development and test sets related to the given set of texts $\mathcal{X}$. $f : \mathcal{X} \rightarrow \mathbb{R}^d$ is the backbone of a language model classifier fine-tuned on $\mathcal{T}$, where $d$ denotes the dimension of $f$ embedding space. 

As shown in Figure~\ref{fig:global_flow}, \method\ is a 4-step method that follows an iterative process, beginning with a concept base covering the text corpus of interest and progressively incorporating the most relevant concepts into the CBL until meeting a completeness stopping criterion, as detailed in Sections~\ref{sec::concept_discovery} to~\ref{sec::stop_criterion}:\\
\textbf{1. Concept Bank Construction}. We ge\-ne\-ra\-te candidate concepts by prompting an auto-regressive SLM to identify micro-concepts (topics) within $\mathcal{T}$.  These micro-concepts are then clustered to form a set of high-level macro-concept candidates $\mathcal{C}$.\\
\textbf{2. Concept Scoring and CBL Initialization}. Each candidate concept in $\mathcal{C}$ receives a score based on its importance for classification and its identifiability in $f$ embedding space. These scores determine which concepts will be incorporated into the concept bottleneck layer. The TCBM CBL is initialized (1) to ensure that nearly every text in the corpus of interest activates at least one concept and (2) to foster concept diversity in the CBL.\\
\textbf{3. TCBM Training}.  Given a set of selected concepts,
%(e.g. as constructed in the first step), 
we train two TCBM variants: a \textit{simple} version using only the explicit concepts and a \textit{residual} one with an additional parallel residual connection capturing additional non interpretable information.\\
\textbf{4. Stopping Criterion}. The training process stops when the performance of the \textit{simple} TCBM achieves comparable performance to the \textit{residual} TCBM, indicating that the explicit concepts from the CBL alone provide a complete basis for the classification task without requiring residual information.

\subsection{Concept Bank Construction}
\label{sec::concept_discovery}

In case of absence of labeled concepts, the first step aims at automatically constructing a set of concept candidates $\mathcal{C}$ without human annotation for potential inclusion in the CBL of the TCBM.

\paragraph{Micro Concept Bank Creation.}
We use a scalable and computationally affordable small language model (9B parameters following ~\citet{slm_survey}) to annotate each text with topic-level "micro concepts" that represent higher-level abstractions than tokens. This creates a micro concept bank~$\widetilde{\mathcal{C}}$ from the text corpus. We give more information about the prompt used to generate micro concepts in Appendix~\ref{sec:prompt_micro_concept}.

\paragraph{Macro Concept Bank Creation.}
We cluster the micro concepts into $p$ macro concepts with $p \ll  \lvert \widetilde{\mathcal{C}} \rvert$, sequentially using sentence embeddings, UMAP~\cite{umap}, and HDBSCAN~\cite{hdbscan}. This addresses semantic redundancy (e.g.,  grouping "demoniac monster" and "diabolic creature" into "supernatural entities"). The final corpus is formalized as $\mathcal{T}_{M} = \left\{ (x,y, \textbf{c})\right\}$, where 
$\textbf{c}  \in  \left\{ 0,1\right\}^{p}$ 
% $\textbf{c} = [c_{1},...,c_{p}] \in  \left\{ 0,1\right\}^{p}$ 
is a vector of absence or presence of the $p$ found macro concepts.

\paragraph{Macro Concept Labeling.}
Each macro concept~$c$ receives a textual label~$l(c)$ by prompting the SLM to identify the superclass of the $m$ micro concepts closest to the macro concept's centroid, with $m=15$ by default.  More details about the prompt used perform macro concept labeling are provided in Appendix~\ref{sec:prompt_macro_concept}.

Thus, unlike its competitors, \method\ offers to perform concept discovery without using ChatGPT and without relying on human annotations.

\subsection{Concept Scoring and CBL Initialization}
Given any labeled concept candidate bank~$\mathcal{C}$ (obtained whether through \method\ concept annotation as in Section~\ref{sec::concept_discovery}, or alternatives), our objective is to determine which subset $\mathcal{C^{*}} \subset \mathcal{C}$ should be included in the CBL. 
%In the following, we call a concept either a macro concept obtained with our annotation approach or a concept obtained in another way. 
\method\ scores each concept $c$ based on linear representations from $f$ embedding and determined by combining two key components: (1) a concept importance measure $i(c)$ and (2) a linear identifiability score $g(c)$. The CBL of the TCBM is then initialized based on these scores and by favoring diversity in the concepts.

\paragraph{Concept Activation Vectors Computation.} 
The "Linear Representation Hypothesis" states that high-level concepts are represented linearly in the embedding space of language models~\cite{elhage_linear_representation,linear_rep_hypothesis}. Motivated by this hypothesis, we assign to each concept~$c$ a linear representation from $f$ embedding space, called Concept Activation Vector (CAV) $\overrightarrow{\gamma(c)}$.  Among the different ways to compute a CAV~\cite{cav_axbench}, we consider the mean difference of embeddings~\cite{steering_mean_embedding} that has been shown to lead to the best compromise is terms of concept detection accuracy and computational cost~\cite{geometry_of_truth}: 
\begin{equation*}
  \overrightarrow{\gamma(c)} = \frac{1}{\left|\mathcal{X}^{c+}_{tr}\right|} \sum_{x\in \mathcal{X}^{c+}_{tr}}{f(x)} - \frac{1}{\left|\mathcal{X}^{c-}_{tr}\right|} \sum_{x \in \mathcal{X}^{c-}_{tr}} {f(x)} 
\end{equation*}
where $\mathcal{X}^{c+}_{tr}$ and $\mathcal{X}^{c-}_{tr}$ respectively represent the sets of texts from the training set~$\mathcal{X}_{train}$ where the concept $c$ is present or absent.

\paragraph{Concept Importance.}
\method\ identifies concepts with high discriminative power by computing their importance. We propose a novel approach, called  \texttt{Concept Integrated Gradients} (\texttt{CIG}),  to estimate the latter, based on \texttt{Integrated gradients}~\cite{sundararajan_axiomatic_2017}. For each input text $x \in \mathcal{X}$ and concept~$c$,  we calculate the local concept importance using the absolute dot product: $\lvert \langle \overrightarrow{\gamma(c)}, \texttt{IG}(f(x))\rangle \rvert$, where $\texttt{IG}(f(x))$ is the neuron importance vector derived by applying \texttt{Integrated Gradients} to $f$ final layer with respect to the ground truth label. This way, the local concept importance is defined as the absolute value of the projection of the $f$ embedding local neuron importance of input $x$ onto the concept direction $\overrightarrow{\gamma(c)}$. The final importance score $i(c)$ is the average of the local concept importance values across $\mathcal{X}_{train}$. \method\ also compute concept importance by appling \texttt{TCAV} as in~\citet{tcav_abusive_language}.  Both approaches identify concepts that play a significant role in $f$ decision-making process, making them valuable candidates for inclusion in the CBL. Further implementation details for both \texttt{CIG} and \texttt{TCAV} are provided in Appendix~\ref{sec:concept_importance}.

\paragraph{Concept Linear Identifiability.}
\method\ identifies concepts that can effectively be detected in $f$ embedding space by computing a linear identifiability score for each concept $c$. For each input text $x \in \mathcal{X}_{dev}$, we predict concept presence based on previously computed CAVs using a simple linear rule:
\begin{align}
\widehat{c}(x) = 
\begin{cases} 
1 & \text{if } \langle \overrightarrow{\gamma(c)}, f(x)\rangle > M_c \\
0 & \text{otherwise}
\end{cases}
\label{eq:concept_linear_predictor}
\end{align}
where $\langle \overrightarrow{\gamma(c)}, f(x)\rangle$ is the projection of the $f$ embedding of input $x$ onto the concept direction $\overrightarrow{\gamma(c)}$. The threshold $M_c$ is the median projection value across~$\mathcal{X}_{dev}$. We then compute the linear identifiability score $g(c)$ as the F1 score between the predictions $\widehat{c}(x)$ and ground truth labels on~$\mathcal{X}_{dev}$. This identifiability score quantifies how accurately each concept can be linearly detected in $f$ embedding space. We define the final concept score as ${i(c) \times g(c)}$, ensuring that selected concepts are both discriminative for the task and reliably detectable from $f$ representations.
%The final concept score for each concept is obtained by multiplying its importance measure $i(c)$ with its linear identifiability sore $g(c)$, ensuring that selected concepts are both discriminative for the task and reliably detectable from $f$ representations.

\paragraph{CBL Initialization.}
We initialize the CBL before the first TCBM training with a minimal set of concepts that covers at least 99\% of $\mathcal{X}_{train}$, ensuring nearly every text activates at least one concept. To achieve this efficiently, we (1) cluster concepts based on their co-occurrence patterns in texts and using HDBSCAN, (2) sort concepts by their previously calculated scores and (3) select concepts iteratively by cluster in descending score order. This approach balances concept relevance with diversity, avoiding redundancy by prioritizing concepts from different clusters. Subsequently, during iterative TCBM training, a concept per cluster is added to the CBL. 

These concept clusters should not be confused with the micro-concept clusters introduced in Section~\ref{sec::concept_discovery}. During the TCBM initialization stage, concept clusters are groups of concepts that activate on the same texts, whereas micro-concept clusters are groups of concepts with similar semantics. In Section~\ref{result}, we compare the clustering-based selection method against a simpler approach that selects concepts solely based on their scores.

\subsection{TCBM Training}

Given a subset of concepts $\text{C} \subset \mathcal{C}$, we introduce the protocol followed by \method\ to train a TCBM. We guide the evolution of the TCBM training by adding a residual connection parallel to the CBL. The residual connection serves two purposes: it acts as an indicator of concept base completeness (see Section~\ref{sec::stop_criterion}) while simultaneously reducing downstream classification leakage. Generating a \textit{simple} TCBM consists in training two layers, $\Phi^{\text{C}}: \mathbb{R}^d \rightarrow \mathbb{R}^{\lvert \text{C} \rvert}$ and $\Phi^{\text{cls}}: \mathbb{R}^{\lvert \text{C} \rvert} \rightarrow \mathcal{Y}$.  $\Phi^{\text{C}}$ is the layer detecting concepts from $f$ embedding and $\Phi^{\text{cls}}$ is the sparse linear concept-based classification layer. The \textit{simple} TCBM is then defined as $\Phi^{\text{cls}}~\circ~\Phi^{\text{C}}~\circ~f$. A \textit{residual} TCBM contains an additional non interpretable residual layer $\Phi^{r}: \mathbb{R}^d \rightarrow \mathcal{Y}$ using unknown residual concepts to enhance the downstream classification accuracy of the TCBM and mitigate leakage. This way, the \textit{residual} TCBM is defined as $((\Phi^{\text{cls}}~\circ~\Phi^{\text{C}})+\Phi^{r})~\circ f$.

\method\ constructs $\Phi^{\text{C}}$ based on supervised learning,
%. The aim is to minimise 
minimising the following loss function: 
\begin{eqnarray}
\label{eqn:argmin2}
  \mathcal{L}_{TCBM} = \lambda \mathcal{L}(\Phi^{\text{C}}(f(x)), \texttt{c})
\end{eqnarray}
$$
  +  \mathcal{L}(\Phi^{\text{cls}}(\Phi^{\text{C}}(f(x)) + \Phi^{r}(f(x)), y)
$$
where $\mathcal{L}$ is the cross-entropy loss, $\lambda$ a hyperparameter and $\texttt{c}$ the vector of absence or presence of the concepts included in \text{C},
i.e. the restriction to~$\text{C}$ of $\textbf{c}$ defined on~$\mathcal C$  in Section~\ref{sec::concept_discovery}. $\Phi^{\text{r}}$ is trained with a ridge penalty constraint as in \citet{cbm_posthoc} and $\Phi^{\text{cls}}$ is trained with an elastic net penalty constraint to foster sparsity. The supervised training can be done \textit{jointly} (concept detection and downstream classification performed at the same time) or \textit{sequentially} (concept detection learned first and classification training performed afterwards). 

% \method\ also implements a TCBM  building method based on the CAVs projection in the concept layer. We present this \texttt{projection} methodology and give more details about training strategies, Ridge and elastic net hyperparameters in Appendix~\ref{sec:appendix_tcbm_implementation_details}.

\subsection{Stopping Criterion}
\label{sec::stop_criterion}
\method\ employs a performance-based criterion to determine when sufficient concepts have been incorporated into the CBL. At each iteration, we evaluate the two variants of our TCBM, the simple and the residual ones.  Indeed, the residual layer $\Phi^{r}$ captures information not yet represented by 
the current concept base. We then approximate concept base completeness through a quantitative performance comparison: 
when the TCBM without~$\Phi^{r}$ 
achieves at least ($1-\epsilon$)\% times the performance of the 
TCBM with $\Phi^{r}$, we conclude that the CBL adequately captures the essential information for classification. This criterion is reasonable because the concept base is initially constructed to cover nearly all texts in $\mathcal{X}$. Therefore, when the TCBM without $\Phi^{r}$ approximates the performance with $\Phi^{r}$, it suggests that the current concepts capture
%our explicit concepts are capturing
the necessary information for the classification task, with minimal reliance on unexplained residual concepts. Once this performance threshold is met, the training process terminates, and we remove the residual layer $\Phi^{r}$ from the model. This final step yields a pure TCBM without any non-interpretable components, ensuring that all classification decisions are solely based on 
%our
the human-understandable CBL.

\begin{table*}[]
% \footnotesize
% \scriptsize
\small
\begin{center}
\begin{tabular}{cc|cccc|cccc}
\hline
\multicolumn{2}{c|}{\textbf{\begin{tabular}[c]{@{}c@{}}Model backbone  (size)\end{tabular}}}                                      & \multicolumn{4}{c|}{\textbf{\begin{tabular}[c]{@{}c@{}}\texttt{BERT-base} (110M)\end{tabular}}}                                                                                                 & \multicolumn{4}{c}{\textbf{\begin{tabular}[c]{@{}c@{}}\texttt{DeBERTa-large}  (395M)\end{tabular}}}                                                                                              \\ \hline
\multicolumn{1}{c|}{\textbf{Dataset}}                                                                      & \textbf{Method}        & \begin{tabular}[c]{@{}c@{}}\texttt{Black}\\ \texttt{box}\end{tabular} & \texttt{C\textsuperscript{3}M}        & \texttt{CB-LLM} & \begin{tabular}[c]{@{}c@{}}\texttt{CT-CBM}\\ (ours)\end{tabular} & \begin{tabular}[c]{@{}c@{}}\texttt{Black}\\ \texttt{box}\end{tabular} & \texttt{C\textsuperscript{3}M}        & \texttt{CB-LLM} & \begin{tabular}[c]{@{}c@{}}\texttt{CT-CBM}\\ (ours)\end{tabular} \\ \hline
\multicolumn{1}{c|}{}                                                                                      & \%ACC \tiny {$\uparrow$} & 91.0                                                                  & \cellcolor[HTML]{EFEFEF}{\ul 90.1}    & 90.0            & \cellcolor[HTML]{EFEFEF}\textbf{90.6}                            & 92.0                                                                  & \cellcolor[HTML]{EFEFEF}{\ul 91.5}    & 90.1            & \cellcolor[HTML]{EFEFEF}\textbf{91.6}                            \\
\multicolumn{1}{c|}{}                                                                                      & \%c \tiny {$\uparrow$}   & -                                                                     & \cellcolor[HTML]{EFEFEF}{\ul 68.8}    & 56.0            & \cellcolor[HTML]{EFEFEF}\textbf{74.5}                            & -                                                                     & \cellcolor[HTML]{EFEFEF}{\ul 80.2}    & 69.2            & \cellcolor[HTML]{EFEFEF}\textbf{88.7}                            \\
\multicolumn{1}{c|}{\multirow{-3}{*}{\textbf{AGNews}}}                                                     & \#c \tiny {$\downarrow$}  & -                                                                     & \cellcolor[HTML]{EFEFEF}{\ul 41}      & {\ul 41}        & \cellcolor[HTML]{EFEFEF}\textbf{12}                              & -                                                                     & \cellcolor[HTML]{EFEFEF}{\ul 41}      & {\ul 41}        & \cellcolor[HTML]{EFEFEF}\textbf{13}                              \\ \hline
\multicolumn{1}{c|}{}                                                                                      & \%ACC \tiny {$\uparrow$} & 99.4                                                                  & \cellcolor[HTML]{EFEFEF}\textbf{99.5} & {\ul 99.3}      & \cellcolor[HTML]{EFEFEF}\textbf{99.5}                            & 99.4                                                                  & \cellcolor[HTML]{EFEFEF}\textbf{99.5} & {\ul 99.4}      & \cellcolor[HTML]{EFEFEF}\textbf{99.5}                            \\
\multicolumn{1}{c|}{}                                                                                      & \%c \tiny {$\uparrow$}   & -                                                                     & \cellcolor[HTML]{EFEFEF}{\ul 79.5}    & 56.0            & \cellcolor[HTML]{EFEFEF}\textbf{92.9}                            & -                                                                     & \cellcolor[HTML]{EFEFEF}{\ul 86.0}    & 42.1            & \cellcolor[HTML]{EFEFEF}\textbf{99.3}                            \\
\multicolumn{1}{c|}{\multirow{-3}{*}{\textbf{DBpedia}}}                                                    & \#c \tiny {$\downarrow$} & -                                                                     & \cellcolor[HTML]{EFEFEF}{\ul 63}      & {\ul 63}        & \cellcolor[HTML]{EFEFEF}\textbf{18}                              & -                                                                     & \cellcolor[HTML]{EFEFEF}{\ul 63}      & {\ul 63}        & \cellcolor[HTML]{EFEFEF}\textbf{19}                              \\ \hline
\multicolumn{1}{c|}{}                                                                                      & \%ACC \tiny {$\uparrow$} & 95.6                                                                  & \cellcolor[HTML]{EFEFEF}\textbf{97.0} & 95.1            & \cellcolor[HTML]{EFEFEF}{\ul 96.1}                               & 95.9                                                                  & \cellcolor[HTML]{EFEFEF}\textbf{97.0} & 97.2            & \cellcolor[HTML]{EFEFEF}{\ul 96.8}                               \\
\multicolumn{1}{c|}{}                                                                                      & \%c \tiny {$\uparrow$}   & -                                                                     & \cellcolor[HTML]{EFEFEF}{\ul 64.7}    & 47.7            & \cellcolor[HTML]{EFEFEF}\textbf{80.2}                            & -                                                                     & \cellcolor[HTML]{EFEFEF}{\ul 75.8}    & 63.4            & \cellcolor[HTML]{EFEFEF}\textbf{86.8}                            \\
\multicolumn{1}{c|}{\multirow{-3}{*}{\textbf{Ledgar}}}                                                     & \#c \tiny {$\downarrow$} & -                                                                     & \cellcolor[HTML]{EFEFEF}{\ul 78}      & {\ul 78}        & \cellcolor[HTML]{EFEFEF}\textbf{13}                              & -                                                                     & \cellcolor[HTML]{EFEFEF}{\ul 78}      & {\ul 78}        & \cellcolor[HTML]{EFEFEF}\textbf{12}                              \\ \hline
\multicolumn{1}{c|}{}                                                                                      & \%ACC \tiny {$\uparrow$} & 62.7                                                                  & \cellcolor[HTML]{EFEFEF}{\ul 56.7}    & \textbf{57.9}   & \cellcolor[HTML]{EFEFEF}{\ul 56.7}                               & 62.6                                                                  & \cellcolor[HTML]{EFEFEF}\textbf{60.0} & {\ul 59.1}      & \cellcolor[HTML]{EFEFEF}\textbf{60.0}                            \\
\multicolumn{1}{c|}{}                                                                                      & \%c \tiny {$\uparrow$}   & -                                                                     & \cellcolor[HTML]{EFEFEF}53.5          & 29.2            & \cellcolor[HTML]{EFEFEF}\textbf{62.3}                            & -                                                                     & \cellcolor[HTML]{EFEFEF}{\ul 57.4}    & 25.2            & \cellcolor[HTML]{EFEFEF}\textbf{76.5}                            \\
\multicolumn{1}{c|}{\multirow{-3}{*}{\textbf{\begin{tabular}[c]{@{}c@{}}Medical\\ Abstract\end{tabular}}}} & \#c \tiny {$\downarrow$} & -                                                                     & \cellcolor[HTML]{EFEFEF}{\ul 57}      & {\ul 57}        & \cellcolor[HTML]{EFEFEF}\textbf{15}                              & -                                                                     & \cellcolor[HTML]{EFEFEF}{\ul 57}      & {\ul 57}        & \cellcolor[HTML]{EFEFEF}\textbf{16}                              \\ \hline
\multicolumn{1}{c|}{}                                                                                      & \%ACC \tiny {$\uparrow$} & 91.7                                                                  & \cellcolor[HTML]{EFEFEF}{\ul 91.6}    & 91.4            & \cellcolor[HTML]{EFEFEF}\textbf{92.7}                            & 93.8                                                                  & \cellcolor[HTML]{EFEFEF}\textbf{93.8} & 90.9            & \cellcolor[HTML]{EFEFEF}{\ul 93.5}                               \\
\multicolumn{1}{c|}{}                                                                                      & \%c \tiny {$\uparrow$}   & -                                                                     & \cellcolor[HTML]{EFEFEF}{\ul 68.7}    & 29.8            & \cellcolor[HTML]{EFEFEF}\textbf{84.5}                            & -                                                                     & \cellcolor[HTML]{EFEFEF}{\ul 77.1}    & 52.2            & \cellcolor[HTML]{EFEFEF}\textbf{84.3}                            \\
\multicolumn{1}{c|}{\multirow{-3}{*}{\textbf{\begin{tabular}[c]{@{}c@{}}Movie\\ Genre\end{tabular}}}}      & \#c \tiny {$\downarrow$} & -                                                                     & \cellcolor[HTML]{EFEFEF}68            & 68              & \cellcolor[HTML]{EFEFEF}\textbf{14}                              & -                                                                     & \cellcolor[HTML]{EFEFEF}{\ul 68}      & {\ul 68}        & \cellcolor[HTML]{EFEFEF}\textbf{13}                              \\ \hline
\end{tabular}
\caption{\label{tab:results} 
\method\ and competitors evaluation on 5 test sets and 2 NLP classifiers. Except the black box baseline, the best results (resp. second best) are in bold (resp. underlined). \method\ and \texttt{C\textsuperscript{3}M} training are based on \texttt{C\textsuperscript{3}M} annotation.}
\end{center}
\end{table*}

\section{Experimental Settings}
\label{xp}

This section presents the experimental study conducted across 5 datasets and 3 NLP classifiers of different sizes. We compare \method\ to several competitors, and we run an ablation study to assess the impact of (1) the concept clustering during the CBL initialization, (2) the method used to compute concept importance and (3) the concept identifiability score. Then we illustrate TCBM applications, namely  concept intervention during inference, better understanding of counterfactual explanations and adversarial attacks~\cite{XAI_nlp_survey} and providing global explanations.

\begin{table}[t]
% \footnotesize
\small
% \footnotesize
\begin{center}
\begin{tabular}{P{0.7cm} P{0.95cm}|P{0.65cm} P{0.82cm}|P{0.65cm} P{0.82cm}}
% \begin{tabular}{cc|cc|cc}
\hline
\multicolumn{2}{c|}{\textbf{\begin{tabular}[c]{@{}c@{}}Concept\\ annotation\end{tabular}}}                                     & \multicolumn{2}{c|}{\textbf{\begin{tabular}[c]{@{}c@{}}\texttt{CT-CBM}\\\end{tabular}}}       & \multicolumn{2}{c}{\textbf{\begin{tabular}[c]{@{}c@{}}\texttt{CT-CBM} + \\ \texttt{C3M}\end{tabular}}} \\ \hline
\multicolumn{1}{c|}{\textbf{Dataset}}                                                                 & \textbf{Method}        & \texttt{C\textsuperscript{3}M}        & \begin{tabular}[c]{@{}c@{}}\texttt{CT-CBM}\\ (ours)\end{tabular} & \texttt{C\textsuperscript{3}M}            & \begin{tabular}[c]{@{}c@{}}\texttt{CT-CBM}\\ (ours)\end{tabular}     \\ \hline
\multicolumn{1}{c|}{}                                                                                 & \%ACC\tiny {$\uparrow$} & \cellcolor[HTML]{EFEFEF}91.1          & \textbf{91.1}                                                    & \cellcolor[HTML]{EFEFEF}90.3              & \textbf{91.1}                                                        \\
\multicolumn{1}{c|}{}                                                                                 & \%c\tiny{$\uparrow$}   & \cellcolor[HTML]{EFEFEF}\textbf{54.8} & 52.1                                                             & \cellcolor[HTML]{EFEFEF}56.2              & \textbf{80.8}                                                        \\
\multicolumn{1}{c|}{\multirow{-3}{*}{\textbf{AGNews}}}                                                & \#c\tiny$\downarrow$  & \cellcolor[HTML]{EFEFEF}100           & \textbf{12}                                                      & \cellcolor[HTML]{EFEFEF}141               & \textbf{12}                                                          \\ \hline
\multicolumn{1}{c|}{}                                                                                 & \%ACC\tiny {$\uparrow$} & \cellcolor[HTML]{EFEFEF}\textbf{99.5} & \textbf{99.5}                                                    & \cellcolor[HTML]{EFEFEF}\textbf{99.5}     & \textbf{99.5}                                                        \\
\multicolumn{1}{c|}{}                                                                                 & \%c\tiny{$\uparrow$}   & \cellcolor[HTML]{EFEFEF}52.0          & \textbf{81.1}                                                    & \cellcolor[HTML]{EFEFEF}58.9              & \textbf{91.3}                                                        \\
\multicolumn{1}{c|}{\multirow{-3}{*}{\textbf{DBpedia}}}                                               & \#c \tiny {$\downarrow$} & \cellcolor[HTML]{EFEFEF}100           & \textbf{18}                                                      & \cellcolor[HTML]{EFEFEF}163               & \textbf{18}                                                          \\ \hline
\multicolumn{1}{c|}{}                                                                                 & \%ACC\tiny {$\uparrow$} & \cellcolor[HTML]{EFEFEF}91.3          & \textbf{92.6}                                                    & \cellcolor[HTML]{EFEFEF}91.5              & \textbf{91.6}                                                        \\
\multicolumn{1}{c|}{}                                                                                 & \%c\tiny {$\uparrow$}   & \cellcolor[HTML]{EFEFEF}45.3          & \textbf{50.4}                                                    & \cellcolor[HTML]{EFEFEF}54.5              & \textbf{82.6}                                                        \\
\multicolumn{1}{c|}{\multirow{-3}{*}{\textbf{\begin{tabular}[c]{@{}c@{}}Movie\\ Genre\end{tabular}}}} & \#c\tiny{$\downarrow$} & \cellcolor[HTML]{EFEFEF}100           & \textbf{14}                                                      & \cellcolor[HTML]{EFEFEF}168               & \textbf{16}                                                          \\ \hline
\end{tabular}
\caption{\label{tab:results_our_annotation} 
\method\ and \texttt{C\textsuperscript{3}M}  evaluation on three test sets on \texttt{BERT}. Evaluation is done either based on \method\ concept annotation or the union of \method\ and \texttt{C\textsuperscript{3}M} concept annotations. The best result are in bold.}
\end{center}
\end{table}

\subsection{Experimental Protocol}

\paragraph{Datasets and models.} \method\ is tested on five multi-class text classification datasets: AG News~\cite{AG_news}, DBpedia~~\cite{dbpedia}, Movie Genre\footnote{\url{https://www.kaggle.com/competitions/movie-genre-classification/overview}} and critical domain datasets such as the legal dataset Ledgar~\cite{tuggener2020ledgar} ans Medical Abstracts~\cite{health_dataset_nlp}. We apply \method\ on three fine-tuned NLP classifiers of different sizes: \texttt{BERT}~\cite{devlin_bert_2019},  \texttt{DeBERTa-large}~\cite{hedeberta} and \texttt{Gemma-2-2B}. More information about the used language models are provided in Appendix~\ref{sec:appendix_classifiers_dataset_details}.

\paragraph{\method\ and Competitors.}
We run \method\ with the \texttt{Gemma-2-9B} SLM to generate concept candidates. The CBL is constructed by \textit{joint} training and important concepts are targeted and added in the bottleneck layer with either \texttt{CIG} and \texttt{TCAV}. The stopping criterion parameter $\epsilon$ is set a 0.05. We compare \method\ to both \texttt{C\textsuperscript{3}M}~\cite{cbm_plm} and \texttt{CB-LLM}~\cite{cb_llm}. Our aim is to enrich a simple classifier to generate concepts prior to a final prediction. Given that \texttt{TBM}~\cite{cbm_by_design} does not enhance an NLP classifier but rather performs concept detection with GPT4 during inference, we do not include it in the comparative study. Since \method\ and \texttt{C\textsuperscript{3}M} work with binary represented concepts, their training is done either based on \method\ or \texttt{C\textsuperscript{3}M} concept annotation. To ensure comparability and address the ChatGPT annotation non scalability of \texttt{C\textsuperscript{3}M} (complexity proportional to size of the dataset $\times$ number of targeted concepts), we run \texttt{C\textsuperscript{3}M} with \texttt{Gemma-2-9B} as concept annotator. For each approach, concept annotation is done on $\mathcal{X}_{train}$, $\mathcal{X}_{dev}$ and $\mathcal{X}_{test}$, to enable to evaluate concept detection (see next paragraph). Each method is trained with an early stopping strategy applied to $\mathcal{X}_{dev}$. The hyperparameters of the experiments are detailed in Appendices~\ref{sec:tcbm_training} and~\ref{sec:appendix_competitors_implementation_details}.

\paragraph{Evaluation Criteria.}
We propose an evaluation with 3 metrics:  (1) final classification task accuracy (\textbf{\%ACC}), (2) concept detection accuracy (\textbf{\%c}) measured by F1 score to address concept label imbalance, (3) number of concepts (\textbf{\#c}) in the CBL. Due to computational constraints, \texttt{C\textsuperscript{3}M} concept detection evaluation uses \texttt{Gemma-2-9B} instead of ChatGPT and is limited to 4000 texts. \texttt{CB-LLM} concept evaluation is done by discretizing its concept prediction since the \texttt{CB-LLM} framework represents concepts with numerical values.

\subsection{Results}
\label{result}

\paragraph{Global Results.} Table~\ref{tab:results} shows the experimental results obtained by \method\ and its competitors on \texttt{BERT} and \texttt{DeBERTa}, where \texttt{C\textsuperscript{3}M} and \texttt{CT-CBM} are trained based on concepts obtained with the \texttt{C\textsuperscript{3}M} protocol. Overall, \method\ achieves a performance very similar to that of the original black box models and its competitors in terms of downstream task accuracy (\%ACC). More importantly, \method\ achieves by far the best results both in terms of CBL size (up to 54 fewer concepts in \#c) and concept detection accuracy (\%c  improvements of 5.7 to 33.1 points) for all datasets and models. These results are achieved by ensuring a good compromise between concept conciseness and expressiveness, since \method\ constructs a TCBM based on a complete concept base properly covering the text corpus. \method\ performs well both on datasets from general domains (AGnews, DBpedia and Movie Genre) and more technical critical domains (Ledgar and Medical Abstract). We give additional results in Appendix~\ref{sec:appendix_additional_results}, Table~\ref{tab:gemma_2_2B} showing that \method\ works well when applied to the \texttt{Gemma-2-2B} classifier on AGNews, DBpedia and the Movie Genre dataset.

\paragraph{Varying the Concept Annotation Method.}
Table~\ref{tab:results_our_annotation} shows the experimental results obtained by appling \method\ and \texttt{C\textsuperscript{3}M} on a \texttt{BERT} base model. In this table \method\ and \texttt{C\textsuperscript{3}M} are either trained based on \method\ annotation or in a case where concept annotations from both methods are available. Except for the AGnews dataset, \method\ always over-performs \texttt{C\textsuperscript{3}M} in terms of concept detection accuracy, with drastically less concepts in its CBL. These results show that \method\ is robust to the concept base it is given as input, almost always leading to the best results. While \method\ annotation is computationally less costly than \texttt{C\textsuperscript{3}M} (number of texts vs number of texts $\times$ number of concepts), it results in concepts that are more difficult to classify (e.g. 52.1\%c vs 74.5\%c for \texttt{BERT}/Agnews). This way, the concept annotation method has to be set considering computational budget constraints. We give additional results comparing \method\ and \texttt{C\textsuperscript{3}M} based on \method\ annotation when applied to \texttt{DeBERTa} in Appendix~\ref{sec:appendix_additional_results}, Table~\ref{tab:additional_results}.

\begin{table}[t]
% \footnotesize
\small
% \scriptsize
\begin{center}
\begin{tabular}{c|c|ccc}
\hline
\textbf{Dataset}                                                                 & \textbf{\begin{tabular}[c]{@{}c@{}}\method\ \\ version\end{tabular}}   & \textbf{\begin{tabular}[c]{@{}c@{}}\%ACC \tiny \\ $\uparrow$\end{tabular}} & \textbf{\begin{tabular}[c]{@{}c@{}}\%c \tiny \\ $\uparrow$\end{tabular}} & \textbf{\begin{tabular}[c]{@{}c@{}}\#c \tiny \\ $\downarrow$\end{tabular}} \\ \hline
                                                                                 & \cellcolor[HTML]{EFEFEF}\textbf{CC-\texttt{CIG}-I}                     & \cellcolor[HTML]{EFEFEF}90.6                                               & \cellcolor[HTML]{EFEFEF}74.5                                             & \cellcolor[HTML]{EFEFEF}\textbf{12}                                        \\
                                                                                 & \textbf{$\overline{\text{CC}}$-\texttt{CIG}-I}                         & 90.4                                                                       & \textbf{83.9}                                                            & 19                                                                         \\
                                                                                 & \cellcolor[HTML]{EFEFEF}\textbf{CC-\texttt{CIG}-$\overline{\text{I}}$} & \cellcolor[HTML]{EFEFEF}\textbf{91.2}                                      & \cellcolor[HTML]{EFEFEF}72.5                                             & \cellcolor[HTML]{EFEFEF}14                                                 \\
\multirow{-4}{*}{\textbf{AGnews}}                                                & \textbf{CC-\texttt{TCAV}-I}                                            & 90.6                                                                       & 70.1                                                                     & \textbf{12}                                                                \\ \hline
                                                                                 & \cellcolor[HTML]{EFEFEF}\textbf{CC-\texttt{CIG}-I}                     & \cellcolor[HTML]{EFEFEF}\textbf{99.5}                                      & \cellcolor[HTML]{EFEFEF}\textbf{92.9}                                    & \cellcolor[HTML]{EFEFEF}\textbf{18}                                        \\
                                                                                 & \textbf{$\overline{\text{CC}}$-\texttt{CIG}-I}                         & 99.4                                                                       & 91.1                                                                     & 26                                                                         \\
                                                                                 & \cellcolor[HTML]{EFEFEF}\textbf{CC-\texttt{CIG}-$\overline{\text{I}}$} & \cellcolor[HTML]{EFEFEF}99.3                                               & \cellcolor[HTML]{EFEFEF}91.4                                             & \cellcolor[HTML]{EFEFEF}\textbf{18}                                        \\
\multirow{-4}{*}{\textbf{DBpedia}}                                               & \textbf{CC-\texttt{TCAV}-I}                                            & \textbf{99.5}                                                              & 92.5                                                                     & \textbf{18}                                                                \\ \hline
                                                                                 & \cellcolor[HTML]{EFEFEF}\textbf{CC-\texttt{CIG}-I}                     & \cellcolor[HTML]{EFEFEF}\textbf{92.7}                                      & \cellcolor[HTML]{EFEFEF}\textbf{84.5}                                    & \cellcolor[HTML]{EFEFEF}\textbf{14}                                        \\
                                                                                 & \textbf{$\overline{\text{CC}}$-\texttt{CIG}-I}                         & 92.1                                                                       & 76.2                                                                     & 24                                                                         \\
                                                                                 & \cellcolor[HTML]{EFEFEF}\textbf{CC-\texttt{CIG}-$\overline{\text{I}}$} & \cellcolor[HTML]{EFEFEF}91.2                                               & \cellcolor[HTML]{EFEFEF}76.4                                             & \cellcolor[HTML]{EFEFEF}13                                                 \\
\multirow{-4}{*}{\textbf{\begin{tabular}[c]{@{}c@{}}Movie\\ Genre\end{tabular}}} & \textbf{CC-\texttt{TCAV}-I}                                            & 92.0                                                                       & 75.1                                                                     & \textbf{14}                                                                \\ \hline
\end{tabular}
\caption{\label{tab:concet_score} 
\method\ ablation study of concept clustering (CC), concept importance (either \texttt{CIG} or \texttt{TCAV}) and concept identifiability score (I). $\overline{\text{CC}}$ and $\overline{\text{I}}$ respectively stand for the cases without concept clustering and without concept identifiability computation.}
\end{center}
\end{table}

\paragraph{Ablation Study.}
Table~\ref{tab:concet_score} shows the experimental results of the \method\ ablation study on \texttt{BERT} based on \texttt{C\textsuperscript{3}M} annotation. We vary concept clustering (CC), local concept importance (\texttt{CIG}, \texttt{TCAV}) and identifiability scoring (I). For each dataset, the \method\ basic settings (CC-\texttt{CIG}-I) give the best compromise in terms of downstream task accuracy, concept detection and CBL size. Not clustering concept ($\overline{\text{CC}}$) when initializing the concept base 
%before TCBM training 
results in many more concepts in the CBL (e.g. 26 vs 18 for DBpedia). Not using the identifiability score leads in average to less accurate concept detection (e.g. 76.4 vs 84.5
for Movie Genre). Using \texttt{CIG} gives slightly better results than using \texttt{TCAV}. These results validate the interest of (1) concept clustering to efficiently cover the text corpus and (2) \texttt{CIG} and identifiability scoring for accurate concept detection. We also compare with a \texttt{random} baseline on 10 runs in Appendix~\ref{sec:appendix_additional_results}, Table~\ref{tab:ablation_study_appendix}, highlighting that \texttt{TCAV} and \texttt{CIG} generate better results in terms of concept accuracy than randomly selected concepts.

\subsection{Practical Applications of TCBM}
\paragraph{TCBM Intervention.}
A common application of CBM is to make domain experts modify concept activations at test time to improve the final task accuracy~\cite{cbm_learning_intervene}. We show in Appendix~\ref{sec:appendix_additional_results}, Table~\ref{tab:appendix_intervention} how \method\ concept intervention during inference improves the accuracy, from 92.7\% to 94.0\% on the Movie Genre dataset and 90.6\% to 91.8\% on AGNews.  

\begin{figure}[t]{\centering}
\begin{center}
\includegraphics[scale=0.35]{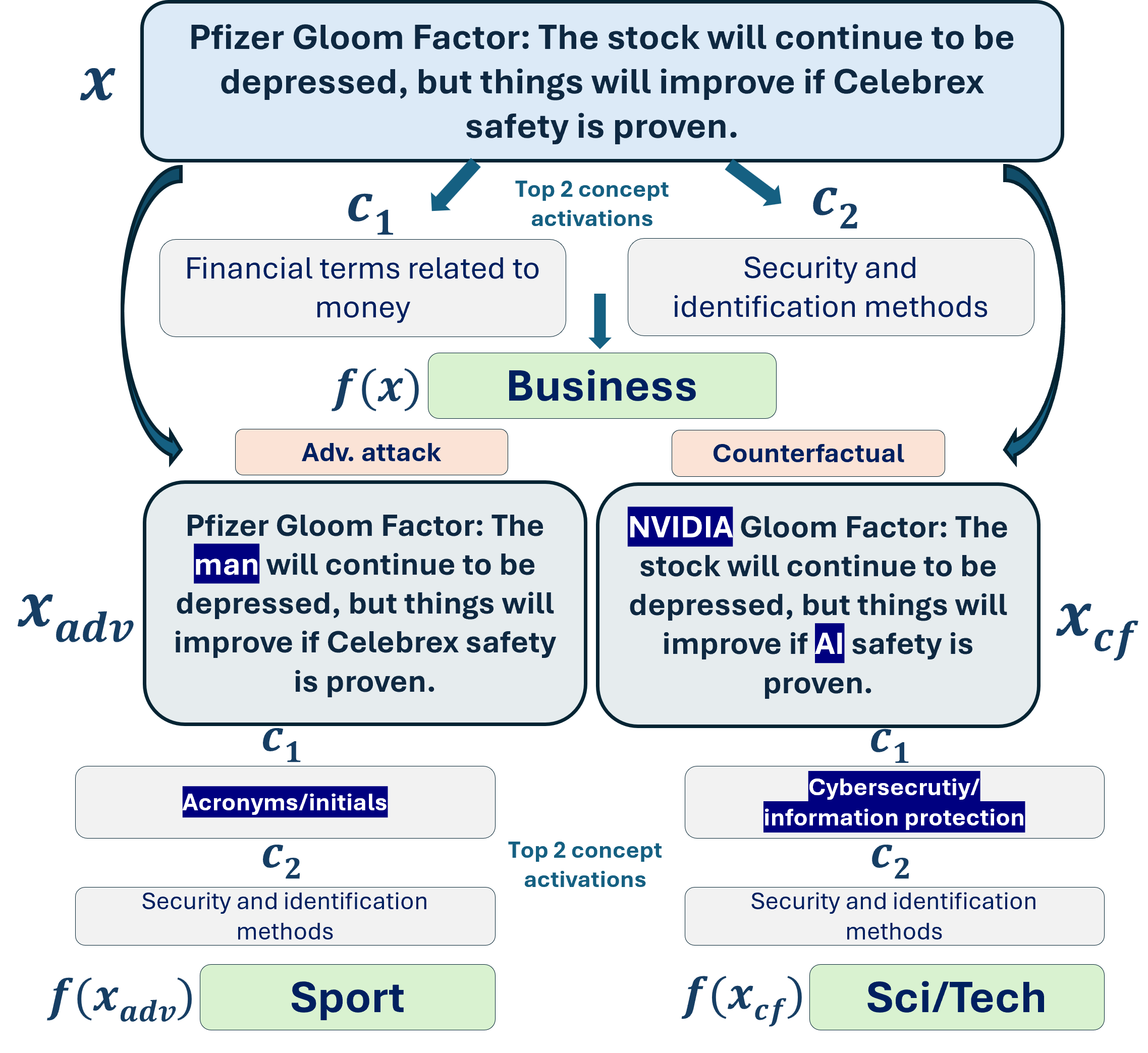}
\caption{Example of an adversarial attack (left, $x_{adv}$) and a counterfactual explanation (right, $x_{cf}$) obtained from \method\ on AGnews. TCBM enables to understand the label change in terms of concept change.}
\label{fig:cf_detox}
\end{center}
\end{figure}

\paragraph{Better Understanding Adversarial Attacks and Counterfactual Explanations.}
We propose to use TCBM for analyzing adversarial attacks and counterfactual explanations by focusing on the concept modifications enabling the prediction switch. As shown in Figure~\ref{fig:cf_detox}, we use \texttt{TextAttack}~\cite{morris2020textattack}  and \texttt{Claude 3.5 Sonnet}, to generate examples that successfully switch the TCBM predictions on the AGnews dataset. An adversarial attack flips a prediction from "Business" to "Sport" by changing the token "stock" to "man". TCBM highlight that this change is interpreted as concept shift from  "Financial terms related to money" to "Acronyms/initials", revealing a concept misunderstanding. Similarly, counterfactual changes from the tokens "Pfizer" to "NVIDIA" and "Celebrex" to "AI" flip the label from "Business" to "Sci/Tech", highlighting a concept shift from "Financial terms" to "Security and identification methods". This concept-level analysis provides deeper understanding than token-level examination alone.

\paragraph{Global TCBM Interpretability.} We finally propose to exploit TCBM at a global scale, by representing the relationship between tokens, concepts and target classes. To identify important tokens for concept activation, we apply Integrated Gradients~\cite{sundararajan_axiomatic_2017} and average the resulting importance scores by concept.  The weights of the~$\Phi^{\text{cls}}$ layers allow to directly represent concept-to-label relationships. This principle is illustrated with Figure~\ref{fig:sankey_t_c_y} in Appendix~\ref{sec::global_interpret_appendix}.

% \method\ on DBPedia can e.g.  show how the "Financial terms related to money" concept significantly influences "Science and Technology" predictions, while the token "internet" strongly activates the "Financial terms related to money" concept.
% %RéessayerClaude peut faire des erreurs. Assurez-vous de vérifier ses réponses.

% Regarding the relationship between important tokens and concepts, we propose to apply an attribution method, here \texttt{Integrated gradients}~\cite{sundararajan_axiomatic_2017} to explain TCBM concept activations for each concept at a global scale. The resulting local feature importance scores are then averaged by concept. Finally, the weights of the~$\Phi^{\text{cls}}$ layers are directly used to represent the intensity of the relationship between  concepts and target labels. Figure~\ref{fig:sankey_t_c_y} gives an example of a global explanation of the proposed \method\ trained on DBPedia. The "Financial terms related to money" concept is important to predict that a press article is related to "Science and Technology". Moreover, "internet" is identified as important to activate the "Financial terms related to money" concept.

\section{Conclusion}
\label{conclusion}
We introduced \method, a novel approach to transform a fine-tuned NLP classifier into a Textual Concept Bottleneck Model. \method\ automatically generates, scores and targets concepts to build a complete Concept Bottleneck Layer. \method\ replicates the same downstream classification performance than its competitors in normal and critical domains dataset, while generating a complete concept base with drastically less concepts, leading to significantly more accurate concept detection. 
Moreover, we highlighted several advantages of TCBM, such as intervening on concepts to improve performance, increasing the intelligibility of adversarial attacks and counterfactuals and producing global explanations. 

\section{Limitations}
\paragraph{Datasets and models.}
This work tested \method\ on five datasets and three language models. It would be interesting to include other models in the study.

\paragraph{Concept Interactions.}
We have not considered possible relationships between concepts. This could highlight a better understanding of the impact of concepts on the classes to be predicted. We see this as a promising way of improving our approach.

\paragraph{Concept Importance.}
There are other approaches for assessing the importance of a concept in explaining the behavior of a model~\cite{fel2023holistic,concept_activation_region}. Using these approaches would enable \method\ to better target important concepts to be added to the CBL.

\paragraph{Text generation.}
Recent work has focused on text generation to generate explanations before answering the question in the same way as TCBM~\cite{bhan-etal-2024-self, cb_llm}. For the time being, our work has focused on text classification.

\label{limitations}

\section*{Ethics Statement}
Since NLP training data can be biased, there is a risk of generating harmful concepts to be added in the CBL. One using \method\ to enhance a NLP classifier must be aware of these biases in order to stand back and analyze the produced concepts and the manipulated texts. Moreover, the use of \texttt{Gemma-9B} for concept annotation is computationally costly and consumes energy, potentially emitting greenhouse gases. \method\ must be  used with caution.

\bibliography{compit}
\bibliographystyle{acl_natbib}
\appendix
\section{Appendix}
\label{sec:appendix}
\subsection{Scientific Libraries}
We used several open-source libraries in this work: pytorch~\cite{paszke2019pytorch}, HuggingFace transformers~\cite{wolf2020transformers} sklearn~\cite{pedregosa2011scikit} and Captum~\cite{miglani_using_2023}. 

\subsection{Autoregressive Language Models Implementation Details}
\label{sec:appendix_slm_implementation_details}
\paragraph{Language Models.} The library used to import the pretrained autoregressive language models is Hugging-Face. In particular, the backbone version of Gemma-2-9B is \texttt{gemma-2-9B-it}.

\paragraph{Gemma-2 Instruction Special Tokens.}
The special tokens to use Gemma in instruction mode were the following:
\begin{itemize}
    \item \texttt{Gemma-2}: \begin{itemize}
        \item \texttt{user\_token= '<start\_of\_turn>user'}
        \item \texttt{assistant\_token= '<start\_of\_turn>model'}
        \item \texttt{stop\_token='<eos>'}
    \end{itemize}
\end{itemize}
\paragraph{Text Generation.}
Text generation was performed using the native functions of the Hugging Face library: \texttt{generate}. The \texttt{generate} function has been used with the following parameters:
\begin{itemize}
    \item \texttt{max\_new\_tokens = 50}
    \item \texttt{do\_sample = True}
    \item \texttt{num\_beams = 2}
    \item \texttt{no\_repeat\_ngram\_size = 2}
    \item \texttt{early\_stopping = True}
    \item \texttt{temperature = 1}
\end{itemize}

\subsection{Prompting Format}
\label{sec:appendix_prompt}
Here we provide some details of different prompts used to give instructions to Gemma-2-9B for micro concept annotation and macro concept labeling. We mainly leverage the In-context Learning (ICL)~\cite{dong_survey_2023}  capabilities of Gemma-2-9B. 
\\

\subsubsection{Preprompt for Micro Concept Generation}
\label{sec:prompt_micro_concept}
\textbf{user}\\
\textit{You are presented with several parts of speech.
        Identify only the main topics in this text. Respond with topic in list format like the examples in a very concise way using as few words as possible. Example: 'As cities expand and populations grow, there is a growing tension between development and the need to preserve historical landmarks. Citizens and authorities often clash over the balance between progress and cultural heritage.'}
\\
\textbf{assistant} \\
\textit{Topics: ['urban development', 'cultural heritage', 'conflict']<eos>}
\\
\textbf{user} \\
\textit{'Recent breakthroughs in neuroscience are shedding light on the complexities of human cognition. Researchers are particularly excited about the potential to better understand decision-making processes and emotional regulation in the brain.'} \\
\textbf{assistant} \\
\textit{Topics: ['neuroscience', 'human cognition', 'decision-making', 'emotional regulation']<eos>}

\subsubsection{Preprompt for Macro Concept Labeling}
\label{sec:prompt_macro_concept}
\textbf{user}\\
\textit{You are presented with several parts of speech.
        Summarise what these parts of speech have in common in a very concise way using as few words as possible. Example: ["piano", "guitar", "saxophone", "violin", "cheyenne", "drum"]}
\\
\textbf{assistant} \\
\textit{Summarization: 'musical instrument'<eos>}
\\
\textbf{user} \\
\textit{["football", "basketball", "baseball", "tennis", "badmington", "soccer"]} \\
\textbf{assistant} \\
\textit{Summarization: 'sport'<eos>} \\
\textbf{user} \\
\textit{["lion", "tiger", "cat", "pumas", "panther", "leopard"]}\\
\textbf{assistant} \\
\textit{Summarization: 'feline-type animal'<eos>}

% \begin{figure*}[t]
% \centering
% \includegraphics[width=0.9\linewidth]{image/generated_text_example.png}
% \caption{ARC Challenge answers conditioned by different ICL prompt built from different rationale generators.}
% \label{fig:text_example}
% \end{figure*}

\subsection{TCBM Implementation Details}
\label{sec:appendix_tcbm_implementation_details}

\subsubsection{Micro Concept Clustering Settings}
\label{sec:micro_concept_clustering}
In order to perform micro concept clustering to build macro concepts, we use the \texttt{umap} library to perform dimension reduction with UMAP with \texttt{n\_components} = 5. Text embeddings are initially obtained with the \texttt{all-mpnet-base-v2} backbone from the \texttt{sentence\_transformers} library. Finally, clustering is performed with \texttt{HDBSCAN} with the basic settings from the \texttt{hdbscan} library.

\subsubsection{Concept Importance Implementation}% Details}
\label{sec:concept_importance}
The \texttt{TCAV} to compute concept-based explanations is done as in~\citet{tcav_abusive_language} by focusing on the final layer of $f$, based on the previously computed CAV $\overrightarrow{\gamma(c)}$. The importance score $i(c)$ is calculated by aggregating the fraction of inputs influenced by each concept with respect to TCAV across all ground truth target classes. Formally, \texttt{TCAV} is locally computed as $\langle \overrightarrow{\gamma(c)}, \nabla f_{cls,k}(f(\mathbf{x}))\rangle$ with $f_{cls,k}$ the classification layer of the initial model, coming after the $f$ backbone, related to the ground truth class~s$k$. In the same way, \texttt{CIG} can be locally formally defined as $\lvert \langle \overrightarrow{\gamma(c)}, \texttt{IG}(f(x))\rangle \rvert$, where $\texttt{IG}(f(x))$ is defined along dimension $i$ as $\texttt{IG}_{i}(f(x)) = (f_i(x) - f_i(x')) \times \int_{0}^{1} (\nabla_{i}f_{cls,k}(f_i(x')+\alpha \times (f_i(x)-f_i(x')) d\alpha)$, with $x'$ a baseline point defined as text with padding only and $f_i(x)$ and $f_i(x')$ the $i$-th neuron of $f(x)$ and $f(x')$.

% \begin{align}
% S_{C,k,l}(\mathbf{x}) &= \langle \overrightarrow{\gamma(c)}, \nabla f_{cls}(f(\mathbf{x}))\rangle,
% \end{align}

\subsubsection{TCBM Training Strategies}
\label{sec:joint_sequential}
\method\ implements two strategies for TCBM training: \textit{joint} and \textit{sequential}. The \textit{sequential} strategy first predicts concepts from input texts and then uses these predicted concepts to make the final target prediction. In this approach, the output of the concept prediction stage is directly used as input for the target prediction stage. This way, the concept loss $\mathcal{L}(\Phi^{\text{C}}(f(x)), \texttt{c})$ is firstly minimized before minimizing the target one $\mathcal{L}(\Phi^{\text{cls}}(\Phi^{\text{C}}(f(x)) + \Phi^{r}(f(x)), y)$. On the other hand, the joint strategy predicts concepts and the final target simultaneously. It optimizes both concept prediction and target prediction losses during training. This enables the model to consider the relationship between concept and target predictions. This way, the loss of Equation~\ref{eqn:argmin2} is directly optimized. In our experiments, TCBMs are trained \textit{jointly} and the $f$ parameters are frozen during the TCBM training.

\subsubsection{Implementation of $\Phi^{r}$ and $\Phi^{\text{cls}}$}
\label{sec:phi_ridge_elasticnet}
$\Phi^{r}$ and $\Phi^{\text{cls}}$ are respectively trained with ridge and elastic net penalties during the TCBM training. The ridge penalization $R$ can be written as follows:
\begin{equation}
\begin{aligned}
    R(W) = \lambda_{R} \|W\|^{2}_2
\end{aligned}
\end{equation}
with $W \in \mathbb{R}^{d \times p}$ the weight matrix of the  $\Phi^{r}$ layer, $\lambda_{R}$ an hyperparameter and $\|\cdot\|^{2}_2$ the $L_2$ norm. 

The elastic net penalization $EN$ can be written as follows:
\begin{equation}
\begin{aligned}
    EN(A) = \lambda_{EN} \left( \alpha \|A\|_1 + (1-\alpha) \|A\|^{2}_2 \right)
\end{aligned}
\end{equation}
with $A  \in \mathbb{R}^{\lvert \text{C} \rvert \times k}$ the weight matrix of the  $\Phi^{\text{cls}}$ layer, $\lambda_{EN}$ and $\alpha$ two hyperparameters and $\|\cdot\|_1$ the $L_1$ norm.

\subsubsection{Other TCBM training hyperparameters}
\label{sec:tcbm_training}
In our experiments, language model and TCBM training is done with the following hyperparameters: \begin{itemize}
    \item \texttt{batch\_size} = 8
    \item \texttt{num\_epochs} = 15
    \item \texttt{max\_len} = 128 for AGnews and DBPedia, 256 for Movie Genre and 512 for Medical Abstracts. 
    \item \texttt{learning rate} = 0.001
    \item \texttt{optimizer} = \texttt{Adam }
    \item $\lambda_{R}$ = 0.01
    \item $\lambda_{EN}$ = 0.5
    \item $\alpha$ = 0.01
    \item $\lambda$ = 0.5
\end{itemize}

\subsubsection{TCBM construction with CAV projection}
\label{sec:cav_projection}
The \texttt{projection} approach to build  $\Phi^{\text{C}}$ consists in projecting the CAVs into the concept space. We formally define $\Phi^{\text{c}_{k}}(f(x)) = \frac{\langle f(x), \overrightarrow{\gamma(c)}\rangle}{||f(x)||.||\overrightarrow{\gamma(c)}||}$ as the linear projection of the embedding of $x$ from $f$ on the concept space associated to concept~$\text{c}$. This way, the concept embedding projection consists in computing the cosine similarity between the CAV and $f$ output. $\Phi^{\text{C}}$ is then constructed by concatenating linear projections corresponding to each concept 
%$\text{c}_{i}$ 
and the final layer. Finally, $\Phi^{\text{cls}}$ and $\Phi^{\text{r}}$ are trained to perform the classification by minimizing the following loss function:
\begin{equation}
 \label{eqn:loss_function}
\mathcal{L}_{TCBM} = \mathcal{L}(\Phi^{\text{cls}}(\Phi^{\text{C}}(f(x)) + \Phi^{r}(f(x)), y)
\end{equation}
where $\mathcal{L}$ is the cross-entropy loss, $\Phi^{r}$ is trained with a ridge penalty constraint and $\Phi^{cls}$ is trained with an elastic net penalty constraint.

\subsection{Language Model Classifiers and Classification Datasets Details}
\label{sec:appendix_classifiers_dataset_details}
\paragraph{Language model classifiers.} The library used to import the pretrained language models is Hugging-Face. In particular, the backbone version of BERT is \texttt{bert-base-uncased} and the one of DeBERTa is \texttt{deberta-large}.

\paragraph{Classification datasets.}
The size of the training sets for AGnews, DBpedia, Movie Genre and Medical Abstracts are respectively 4000, 6000, 4000 and 5000. The size of the test sets for AGnews, DBpedia, Movie Genre and Medical Abstracts are respectively 23778, 30000, 7600 and 2888. \texttt{C\textsuperscript{3}M} concept evaluation is done on 1000 randomly selected rows on each dataset.

\subsection{Competitors Implementation Details}
\label{sec:appendix_competitors_implementation_details}
In our experiments, \texttt{C\textsuperscript{3}M}~\cite{cbm_plm} training is done with the following hyperparameters: \begin{itemize}
    \item \texttt{batch\_size} = 8
    \item \texttt{num\_epochs} = 15
    \item \texttt{max\_len} = 128 for AGnews and DBPedia, 256 for Movie Genre and 512 for Medical Abstracts. 
    \item \texttt{learning rate} = 0.001
    \item \texttt{optimizer} = \texttt{Adam }
    \item $\lambda_{R}$ = 0.01
    \item $\lambda_{EN}$ = 0.5
    \item $\alpha$ = 0.01
    \item $\lambda$ = 0.5
\end{itemize}

The training of \texttt{CB-LLM} is a two-stage process: (1) CBL training and (2) classification layer training. The CBL training is done with the following hyperparameters:  \begin{itemize}
    \item \texttt{batch\_size} = 16
    \item \texttt{num\_epochs} = 4
    \item \texttt{max\_len} = 128 for AGnews and DBPedia, 256 for Movie Genre and 512 for Medical Abstracts. 
    \item \texttt{learning rate} = 0.001
    \item \texttt{optimizer} = \texttt{Adam}
    \item \texttt{loss\_function} = \textit{cos cubed} as in~\citet{label_free_cbm} and ~\citet{cb_llm} 
\end{itemize}

The training of the classification layer of \texttt{CB-LLM} is done with the following hyperparameters: \begin{itemize}
    \item \texttt{batch\_size} = 64
    \item \texttt{num\_epochs} = 50
    \item \texttt{max\_len} = 128 for AGnews and DBPedia, 256 for Movie Genre and 512 for Medical Abstracts. 
    \item \texttt{learning rate} = 0.001
    \item \texttt{optimizer} = \texttt{Adam}
\end{itemize}

\subsection{Post hoc Attribution Explanation Methods}
\label{sec:appendix_post_hoc}
\paragraph{Captum library.}
Post hoc attribution has been computed using the Captum~\cite{miglani_using_2023} library.
In particular, \texttt{Integrated gradients} is computed with respect to language models' embedding layer with Captum's default settings. The embedding layers of BERT and DeBERTa are specified as follows: \texttt{model.model.embed\_tokens}.

%\subsection{\method\ Output Examples}

\label{sec:appendix_examples}
% \subsubsection{Residual connection importance evolution}

% \begin{figure}[H]{\centering}
% \begin{center}
% \includegraphics[scale=0.15]{image/bert_gemma_2b2_20iteration_movies.png}
% \caption{\label{fig:concept_add} Residual connection importance evolution during \method\ BERT training for the movie genre dataset.}
% \end{center}
% \end{figure}

\subsection{Examples of \method\ Macro Concept Compositions}
Figures~3 to~8 illustrates examples of micro-concepts clusters, thus illustrating several macro concepts with their assigned label. 

\begin{figure}[H]{\centering}
\begin{center}
\includegraphics[scale=0.30]{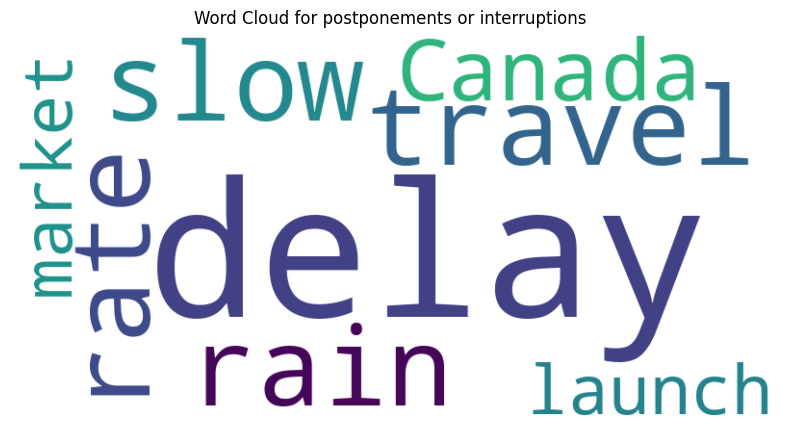}
\caption{\label{fig:concept_add} Cloud of micro concepts composing the macro concept "Postponements or interputions" from the AGnews dataset.}
\end{center}
\end{figure}

\begin{figure}[H]{\centering}
\begin{center}
\includegraphics[scale=0.30]{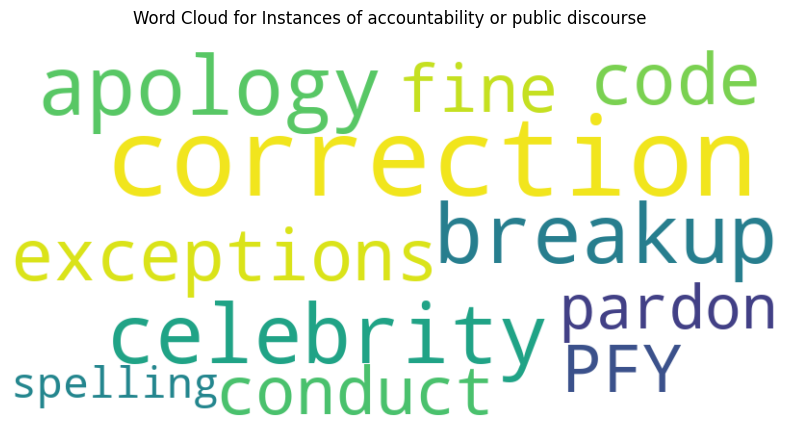}
\caption{\label{fig:concept_add} Cloud of micro concepts composing the macro concept "Instances of accountability or public discourse" from the AGnews dataset.}
\end{center}
\end{figure}

\begin{figure}[H]{\centering}
\begin{center}
\includegraphics[scale=0.30]{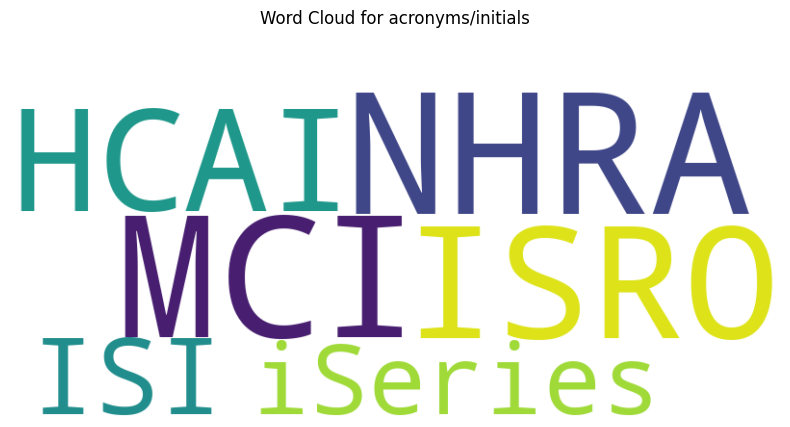}
\caption{\label{fig:concept_add} Cloud of micro concepts composing the macro concept "Acronyms and initials" from the AGnews dataset.}
\end{center}
\end{figure}

\begin{figure}[H]{\centering}
\begin{center}
\includegraphics[scale=0.30]{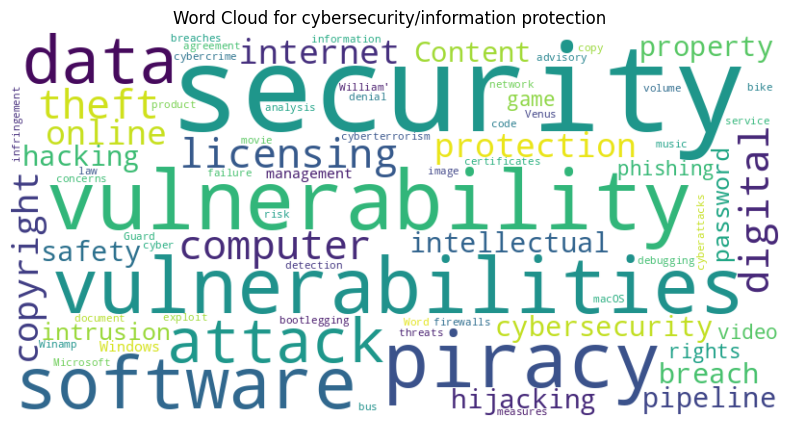}
\caption{\label{fig:concept_add} Cloud of micro concepts composing the macro concept "Cybersecurity and information protection" from the AGnews dataset.}
\end{center}
\end{figure}

\begin{figure}[H]{\centering}
\begin{center}
\includegraphics[scale=0.30]{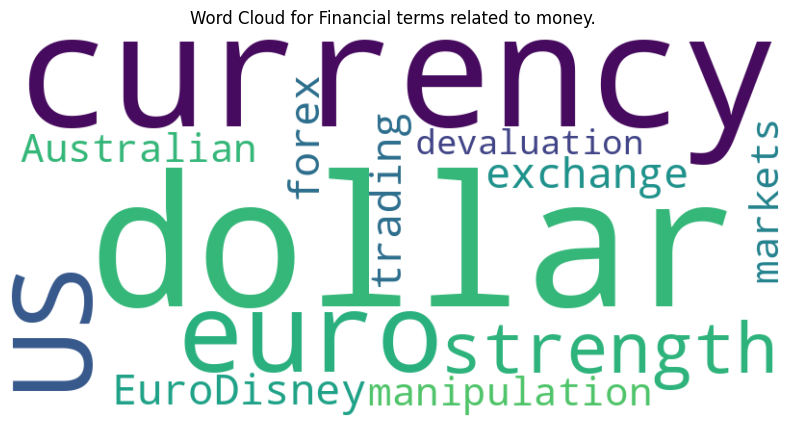}
\caption{\label{fig:concept_add} Cloud of micro concepts composing the macro concept "Financial terms related to money" from the AGnews dataset.}
\end{center}
\end{figure}

\begin{figure}[H]{\centering}
\begin{center}
\includegraphics[scale=0.30]{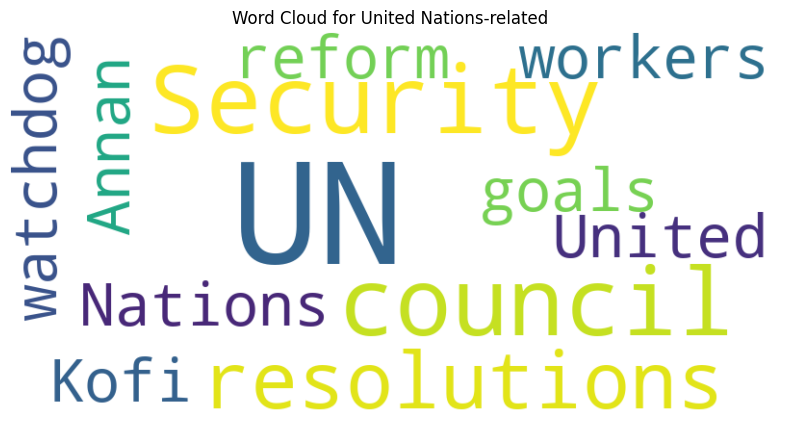}
\caption{\label{fig:concept_add} Cloud of micro concepts composing the macro concept "United-Nations-related" from the AGnews dataset.}
\end{center}
\end{figure}

\subsection{TCBM Global Interpretability}
\label{sec::global_interpret_appendix}

\begin{figure*}[t]{\centering}
\begin{center}
\includegraphics[scale=0.50]{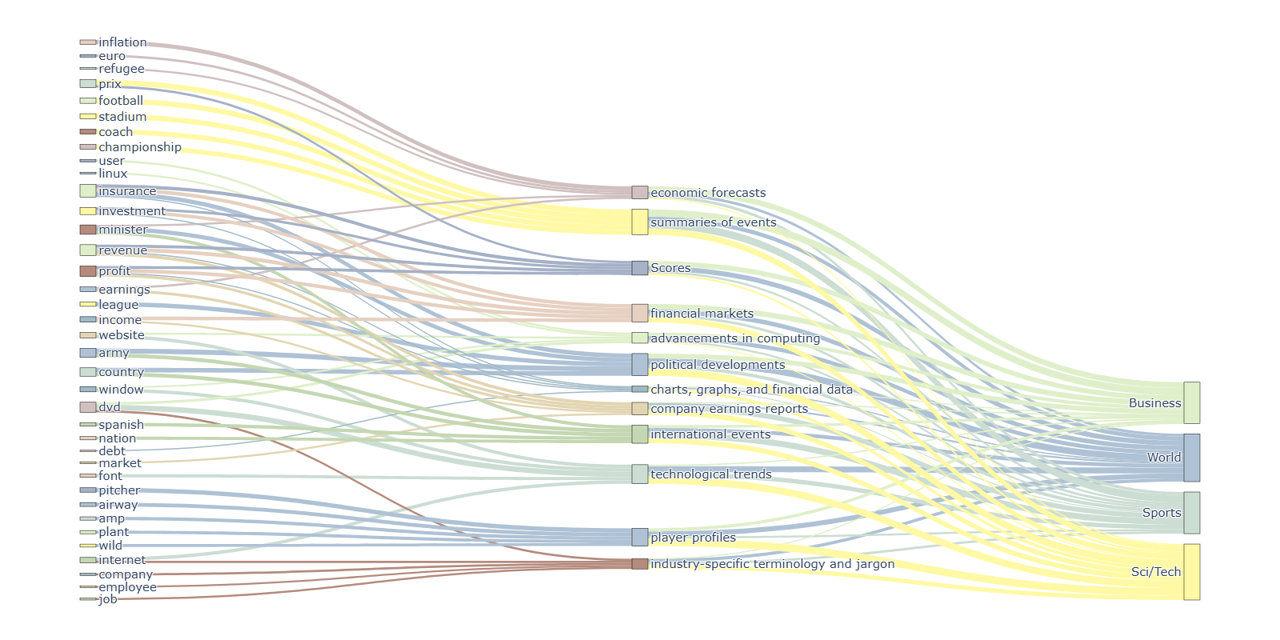}
\caption{Global explanation of a TCBM trained on the AGnews dataset with the \method\ method.}  
\label{fig:sankey_t_c_y}
\end{center}
\end{figure*}

Figure~\ref{fig:sankey_t_c_y} illustrates the proposed use of TCBM for generating global explanations. 

\subsection{Additional Experimental Results}
\label{sec:appendix_additional_results}
In this section we show additional results to highlight the performance of \method\. Table~\ref{tab:additional_results} shows the performance 
of \method\ and \texttt{C\textsuperscript{3}M} when applied to both \texttt{BERT} and \texttt{DeBERTa} on AGnews, DBpedia and Movie Genre, based on the \method\ concept annotation. Table~\ref{tab:ablation_study_appendix} shows the ablation study of \method\ on AGnews, DBpedia and Movie Genre on \texttt{BERT}. Concept clustring (CC), local concept importance (\texttt{CIG}, \texttt{TCAV} and \texttt{random}) and concept identifiability. Results for \texttt{random} are shown by computing the average and the standard deviation on 10 TCBM trainings. 

\begin{table*}[]
\small
\centering
\begin{tabular}{c|c|cc}
\hline
\textbf{Dataset}                                                                & \textbf{Method}        & \textbf{\begin{tabular}[c]{@{}c@{}}Black-box\\ \texttt{Gemma-2-2B}\end{tabular}} & \textbf{\begin{tabular}[c]{@{}c@{}}\method\\\ \texttt{Gemma-2-2B}\end{tabular}} \\ \hline
\multirow{3}{*}{\textbf{AGNews}}                                                & \%ACC \tiny $\uparrow$ & 85.7                                                                             & 91.1                                                                            \\
                                                                                & \%c \tiny $\uparrow$   & -                                                                                & 83.9                                                                            \\
                                                                                & \#c \tiny$\downarrow$  & -                                                                                & 12                                                                              \\ \hline
\multirow{3}{*}{\textbf{DBpedia}}                                               & \%ACC \tiny $\uparrow$ & 95.3                                                                             & 99.2                                                                            \\
                                                                                & \%c \tiny $\uparrow$   & -                                                                                & 99.5                                                                            \\
                                                                                & \#c \tiny $\downarrow$ & -                                                                                & 19                                                                              \\ \hline
\multirow{3}{*}{\textbf{\begin{tabular}[c]{@{}c@{}}Movie\\ Genre\end{tabular}}} & \%ACC \tiny $\uparrow$ & 84.7                                                                             & 89.5                                                                            \\
                                                                                & \%c \tiny $\uparrow$   & -                                                                                & 74.7                                                                            \\
                                                                                & \#c \tiny $\downarrow$ & -                                                                                & 12                                                                              \\ \hline
\end{tabular}
\caption{\label{tab:gemma_2_2B} \method\ performance when applied to the \texttt{Gemma-2-2B} classifier on AGNews, DBPedia and the Movie Genre dataset.}
\end{table*}

\begin{table*}[]
\small
% \scriptsize
\begin{center}
\begin{tabular}{cc|cc|cc}
\hline
\multicolumn{2}{c|}{\textbf{\begin{tabular}[c]{@{}c@{}}Model backbone \\ (size)\end{tabular}}}                                 & \multicolumn{2}{c|}{\textbf{\begin{tabular}[c]{@{}c@{}}\texttt{BERT-base} \\ (110M)\end{tabular}}}       & \multicolumn{2}{c}{\textbf{\begin{tabular}[c]{@{}c@{}}\texttt{DeBERTa-large} \\ (395M)\end{tabular}}}    \\ \hline
\multicolumn{1}{c|}{\textbf{Dataset}}                                                                 & \textbf{Method}        & \texttt{C\textsuperscript{3}M}        & \begin{tabular}[c]{@{}c@{}}\texttt{CT-CBM}\\ (ours)\end{tabular} & \texttt{C\textsuperscript{3}M}        & \begin{tabular}[c]{@{}c@{}}\texttt{CT-CBM}\\ (ours)\end{tabular} \\ \hline
\multicolumn{1}{c|}{}                                                                                 & \%ACC \tiny $\uparrow$ & \cellcolor[HTML]{EFEFEF}91.1          & \textbf{91.1}                                                    & \cellcolor[HTML]{EFEFEF}\textbf{92.1} & 91.2                                                             \\
\multicolumn{1}{c|}{}                                                                                 & \%c \tiny $\uparrow$   & \cellcolor[HTML]{EFEFEF}\textbf{54.8} & 52.1                                                             & \cellcolor[HTML]{EFEFEF}55.0          & \textbf{55.4}                                                    \\
\multicolumn{1}{c|}{}                                                                                 & \#c \tiny$\downarrow$  & \cellcolor[HTML]{EFEFEF}100           & \textbf{12}                                                      & \cellcolor[HTML]{EFEFEF}100           & \textbf{13}                                                      \\
\multicolumn{1}{c|}{\multirow{-4}{*}{\textbf{AGNews}}}                                                & \%D \tiny $\uparrow$   & \cellcolor[HTML]{EFEFEF}78.5          & \textbf{79.3}                                                    & \cellcolor[HTML]{EFEFEF}78.5          & \textbf{79.1}                                                    \\ \hline
\multicolumn{1}{c|}{}                                                                                 & \%ACC \tiny $\uparrow$ & \cellcolor[HTML]{EFEFEF}\textbf{99.5} & \textbf{99.5}                                                    & \cellcolor[HTML]{EFEFEF}\textbf{99.5} & \textbf{99.4}                                                    \\
\multicolumn{1}{c|}{}                                                                                 & \%c \tiny $\uparrow$   & \cellcolor[HTML]{EFEFEF}52.0          & \textbf{81.1}                                                    & \cellcolor[HTML]{EFEFEF}53.0          & \textbf{76.0}                                                    \\
\multicolumn{1}{c|}{}                                                                                 & \#c \tiny $\downarrow$ & \cellcolor[HTML]{EFEFEF}{\ul 100}     & \textbf{18}                                                      & \cellcolor[HTML]{EFEFEF}{\ul 100}     & \textbf{19}                                                      \\
\multicolumn{1}{c|}{\multirow{-4}{*}{\textbf{DBpedia}}}                                               & \%D \tiny $\uparrow$   & \cellcolor[HTML]{EFEFEF}80.3          & \textbf{77.5}                                                    & \cellcolor[HTML]{EFEFEF}\textbf{80.3} & 76.9                                                             \\ \hline
\multicolumn{1}{c|}{}                                                                                 & \%ACC \tiny $\uparrow$ & \cellcolor[HTML]{EFEFEF}91.3          & \textbf{92.6}                                                    & \cellcolor[HTML]{EFEFEF}92.4          & \textbf{92.7}                                                    \\
\multicolumn{1}{c|}{}                                                                                 & \%c \tiny $\uparrow$   & \cellcolor[HTML]{EFEFEF}45.3          & \textbf{50.4}                                                    & \cellcolor[HTML]{EFEFEF}51.5          & \textbf{52.1}                                                    \\
\multicolumn{1}{c|}{}                                                                                 & \#c \tiny $\downarrow$ & \cellcolor[HTML]{EFEFEF}100           & \textbf{14}                                                      & \cellcolor[HTML]{EFEFEF}100           & \textbf{13}                                                      \\
\multicolumn{1}{c|}{\multirow{-4}{*}{\textbf{\begin{tabular}[c]{@{}c@{}}Movie\\ Genre\end{tabular}}}} & \%D \tiny $\uparrow$   & \cellcolor[HTML]{EFEFEF}78.8          & \textbf{78.9}                                                    & \cellcolor[HTML]{EFEFEF}\textbf{78.8} & \textbf{78.1}                                                    \\ \hline
\end{tabular}
\caption{\label{tab:additional_results} 
\method\ and \texttt{C\textsuperscript{3}M} evaluation on three test sets and two NLP classifiers based on the \method\ concept annotation. The best results is highlighted in bold.}
\end{center}
\end{table*}

\begin{table*}[t]
\small
% \scriptsize
\begin{center}
\begin{tabular}{c|c|ccc}
\hline
\textbf{Dataset}                                                                 & \textbf{\begin{tabular}[c]{@{}c@{}}\method\ \\ version\end{tabular}}   & \textbf{\begin{tabular}[c]{@{}c@{}}\%ACC \tiny \\ $\uparrow$\end{tabular}} & \textbf{\begin{tabular}[c]{@{}c@{}}\%c \tiny \\ $\uparrow$\end{tabular}} & \textbf{\begin{tabular}[c]{@{}c@{}}\#c \tiny \\ $\downarrow$\end{tabular}} \\ \hline
                                                                                 & \cellcolor[HTML]{EFEFEF}\textbf{CC-\texttt{CIG}-I}                     & \cellcolor[HTML]{EFEFEF}90.6                                               & \cellcolor[HTML]{EFEFEF}74.5                                             & \cellcolor[HTML]{EFEFEF}\textbf{12}                                        \\
                                                                                 & \textbf{$\overline{\text{CC}}$-\texttt{CIG}-I}                         & 90.4                                                                       & \textbf{83.9}                                                            & 19                                                                         \\
                                                                                 & \cellcolor[HTML]{EFEFEF}\textbf{CC-\texttt{CIG}-$\overline{\text{I}}$} & \cellcolor[HTML]{EFEFEF}\textbf{91.2}                                      & \cellcolor[HTML]{EFEFEF}72.5                                             & \cellcolor[HTML]{EFEFEF}14                                                 \\
                                                                                 & \textbf{CC-\texttt{TCAV}-I}                                            & 90.6                                                                       & 70.1                                                                     & \textbf{12}                                                                \\
\multirow{-5}{*}{\textbf{AGnews}}                                                & \textbf{CC-\texttt{random}-I}                                          & 90.5\tiny{$\pm$0.9}                                                        & 73.9\tiny{$\pm$4.3}                                                      & 13.5\tiny{$\pm$2.1}                                                        \\ \hline
                                                                                 & \cellcolor[HTML]{EFEFEF}\textbf{CC-\texttt{CIG}-I}                     & \cellcolor[HTML]{EFEFEF}\textbf{99.5}                                      & \cellcolor[HTML]{EFEFEF}\textbf{92.9}                                    & \cellcolor[HTML]{EFEFEF}\textbf{18}                                        \\
                                                                                 & \textbf{$\overline{\text{CC}}$-\texttt{CIG}-I}                         & 99.4                                                                       & 91.1                                                                     & 26                                                                         \\
                                                                                 & \cellcolor[HTML]{EFEFEF}\textbf{CC-\texttt{CIG}-$\overline{\text{I}}$} & \cellcolor[HTML]{EFEFEF}99.3                                               & \cellcolor[HTML]{EFEFEF}91.4                                             & \cellcolor[HTML]{EFEFEF}\textbf{18}                                        \\
                                                                                 & \textbf{CC-\texttt{TCAV}-I}                                            & \textbf{99.5}                                                              & 92.5                                                                     & \textbf{18}                                                                \\
\multirow{-5}{*}{\textbf{DBpedia}}                                               & \textbf{CC-\texttt{random}-I}                                          & \textbf{99.5\tiny($\pm$0.3)}                                               & 78.6\tiny{$\pm$11.6}                                                     & 18.9\tiny{$\pm$1.0}                                                        \\ \hline
                                                                                 & \cellcolor[HTML]{EFEFEF}\textbf{CC-\texttt{CIG}-I}                     & \cellcolor[HTML]{EFEFEF}\textbf{92.7}                                      & \cellcolor[HTML]{EFEFEF}\textbf{84.5}                                    & \cellcolor[HTML]{EFEFEF}\textbf{14}                                        \\
                                                                                 & \textbf{$\overline{\text{CC}}$-\texttt{CIG}-I}                         & 92.1                                                                       & 76.2                                                                     & 24                                                                         \\
                                                                                 & \cellcolor[HTML]{EFEFEF}\textbf{CC-\texttt{CIG}-$\overline{\text{I}}$} & \cellcolor[HTML]{EFEFEF}91.2                                               & \cellcolor[HTML]{EFEFEF}76.4                                             & \cellcolor[HTML]{EFEFEF}13                                                 \\
                                                                                 & \textbf{CC-\texttt{TCAV}-I}                                            & 92.0                                                                       & 75.1                                                                     & \textbf{14}                                                                \\
\multirow{-5}{*}{\textbf{\begin{tabular}[c]{@{}c@{}}Movie\\ Genre\end{tabular}}} & \textbf{CC-\texttt{random}-I}                                          & 91.7\tiny{$\pm$0.3}                                                        & 71.9\tiny{$\pm$2.8}                                                      & 14.1\tiny{$\pm$2.0}                                                        \\ \hline
\end{tabular}
\caption{\label{tab:ablation_study_appendix} 
\method\ ablation study of concept clustering (CC), concept importance (either \texttt{CIG}, \texttt{TCAV} or \texttt{random}) and concept identifiability score (I). The \texttt{random} importance score is a score randomly generated, run 10 times to compute the average and the standard deviation. $\overline{\text{CC}}$ and $\overline{\text{I}}$ respectively stand for the cases without concept clustering and without concept identifiability computation during concept scoring.}
\end{center}
\end{table*}

\begin{table*}[]
\small
\centering
\begin{tabular}{c|cc}
\hline
                                                                                             & \multicolumn{2}{c}{\textbf{\begin{tabular}[c]{@{}c@{}}TCBM \\ Performance\end{tabular}}} \\ \cline{2-3} 
\multirow{-2}{*}{\textbf{\begin{tabular}[c]{@{}c@{}}Number of\\ interventions\end{tabular}}} & \textbf{AGnews}     & \textbf{\begin{tabular}[c]{@{}c@{}}Movie\\ Genre\end{tabular}}     \\ \hline
0                                                                                            & 90.6                & 92.7                                                               \\
\rowcolor[HTML]{EFEFEF} 
1                                                                                            & 91.3                & 93.6                                                               \\
2                                                                                            & 91.6                & 93.8                                                               \\
\rowcolor[HTML]{EFEFEF} 
3                                                                                            & 91.7                & \textbf{94.0}                                                      \\
4                                                                                            & \textbf{91.8}       & \textbf{94.0}                                                      \\ \hline
\end{tabular}
\caption{\label{tab:appendix_intervention} Performance of \method\ applied to \texttt{BERT} with respect to the number of interventions during inference on AGnews and the Movie Genre dataset.}
\end{table*}

\end{document}